\documentclass[final
]{elsarticle}
\usepackage{hyperref}
\usepackage{graphicx}
\usepackage{amsmath}
\usepackage{amssymb}
\usepackage{bm}
\usepackage{booktabs}
\usepackage{multirow}
\usepackage{tabularx}
\usepackage[percent]{overpic}
\newcommand{\be}{\begin{equation}}
\newcommand{\ee}{\end{equation}}
\newcommand{\bi}{\begin{itemize}}
\newcommand{\ei}{\end{itemize}}
\newcommand{\bea}{\begin{eqnarray}}
\newcommand{\eea}{\end{eqnarray}}

\newcommand{\bfg}{\mathbf{g}}

\newcommand{\bfx}{\mathbf{x}}

\newcommand{\bfz}{\mathbf{z}}

\newcommand{\cut}[1]{}

\def \ie {i.e.\ }
\usepackage{algorithm,algorithmic}
\usepackage{enumitem}
\usepackage{xspace}
\newcommand{\ourMethodFull}{Learning Direct Optimization\xspace}
\newcommand{\ourMeth}{LiDO\xspace}
\newcommand{\GR}{ground plane\xspace}
\newcommand{\SIM}{Simp\xspace}
\def \eg {e.g.\ }

\newcommand{\etal}{{et al}.\@\xspace}

\usepackage{xcolor}
\definecolor{my2}{RGB}{0, 60, 255}
\definecolor{my3}{RGB}{205,0,30}
\newcommand{\VEC}{\mathbf}

\newcommand{\Red}{\textcolor{black}}

\pdfpxdimen=\dimexpr 1 in/96\relax
\journal{Pattern Recognition}

\makeatletter
\def\blfootnote{\xdef\@thefnmark{}\@footnotetext}
\makeatother

\begin{document}

\begin{frontmatter}

\title{\ourMethodFull for Scene Understanding}

\author[addUoE]{Lukasz Romaszko\corref{mycorrespondingauthor}}
\cortext[mycorrespondingauthor]{Corresponding author}
\ead{lukasz.romaszko@gmail.com}
\author[addUoE,addATI]{Christopher K.I. Williams}
\ead{ckiw@inf.ed.ac.uk}
\author[addMSR]{John Winn}
\ead{jwinn@microsoft.com}



\address[addUoE]{School of Informatics, University of Edinburgh, 10 Crichton St., Edinburgh, EH8 9AB, UK}
\address[addATI]{The Alan Turing Institute, 96 Euston Road, London NW1 2DB, UK}
\address[addMSR]{Microsoft Research, 21 Station Road, Cambridge, CB1 2FB, UK }

\begin{abstract}
  We develop a \ourMethodFull (\ourMeth) method for the refinement of a latent
  variable model that describes input image $\bfx$. Our goal is to
  explain a single image $\bfx$ with an interpretable 3D computer graphics model
  having scene graph latent variables $\bfz$ (such as object
  appearance, camera position).  Given a current estimate of $\bfz$ we
  can render a prediction of the image $\bfg(\bfz)$, which can be
  compared to the image $\bfx$. The standard way to proceed is then to
  measure the error $E(\bfx, \bfg(\bfz))$ between the two, and use an
  optimizer to minimize the error. 
 However, it is unknown which error measure $E$ would be most effective for simultaneously addressing issues such as misaligned objects, occlusions, textures, etc. 
   In contrast, the \ourMeth 
  approach trains a Prediction Network to predict an update directly
  to correct $\bfz$, rather than minimizing the error with respect to
  $\bfz$. Experiments show that  \ourMeth converges rapidly
  as it does not need to perform a search on the error landscape,
  produces better solutions than error-based competitors, and is able
  to handle the mismatch between the data and the fitted scene model.
  We apply \ourMeth to a realistic synthetic dataset, and show
  that the method also transfers to work well with real images.
\end{abstract}


\end{frontmatter}

\section{Introduction}

In this paper we study a \emph{\ourMethodFull} (\ourMeth) method for
the optimization of a latent variable model applied to the problem of
explaining an image $\bfx$ in terms of a computer graphics model.

\blfootnote{Please cite this article as: L. Romaszko, C.K.I. Williams and J. Winn, Learning Direct Optimization for Scene Understanding, Pattern Recognition, Volume 105, 2020, 107369,
\url{https://doi.org/10.1016/j.patcog.2020.107369}}

\begin{figure}[h!t!]

\centering
\includegraphics[width=0.241\textwidth,trim={0cm 0cm 0cm 0cm},clip]{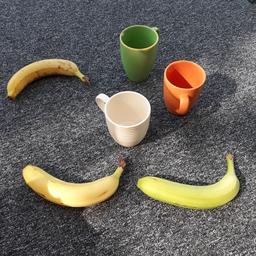}
\includegraphics[width=0.241\textwidth,trim={0cm 0cm 0cm 0cm},clip]{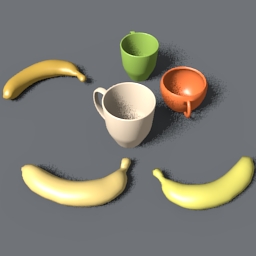}
\includegraphics[width=0.241\textwidth,trim={0cm 0cm 0cm 0cm},clip]{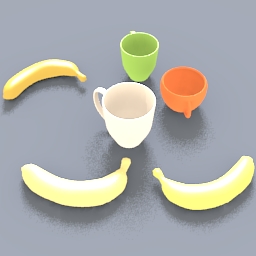}
\includegraphics[width=0.241\textwidth,trim={0cm 0cm 0cm 0cm},clip]{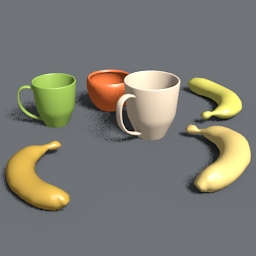}
\newcolumntype{Y}{>{\centering\arraybackslash}X}
\begin{tabularx}{0.99\textwidth}{YYYY}
(a) & (b) & (c) & (d) 
\end{tabularx}
\caption{ (a) Real input image, (b) the rendered image of the 3D scene based on the estimated LVs, (c) the scene under
modified illumination and (d) from a different viewpoint. \label{i:view}}
\end{figure}

The latent variables (LVs) $\bfz$ are the scene graph\footnote{The
  term \emph{scene graph} is taken from the computer graphics
  literature, see e.g.\ \cite{angel-03}.}, i.e.:\ the shape, appearance,
position and poses of all objects in the scene, plus global variables 
such as the camera and lighting.  Our work is carried out in an
analysis-by-synthesis framework. We develop methods to the problem
of scene understanding in 3D from a single image, see Figure~\ref{i:view} --- this is to be
contrasted with methods that simply predict image-based bounding
boxes or pixel labelling.  
Due to the
interpretable 3D representation, one could easily edit the scene, e.g. refine object
positions or their colours, or analyse the 3D scene, e.g.\ compute
possible paths so as not to collide with the objects present in the scene.

Given $\bfz$ we can render the scene graph to
obtain a predicted image $\bfg(\bfz)$, which can be compared to
$\bfx$. The usual way to improve the match is to measure the
error $E(\bfx,\bfg(\bfz))$, and to use an optimizer to update $\bfz$ via minimization of $E(\bfx,\bfg(\bfz))$. 
The \ourMethodFull approach is based on the
idea that we can \emph{train
a network} to predict an update for $\bfz$ rather than
 requiring an error measure $E$ to be defined in the image space, and then minimizing it.

The structure of the paper is as follows: in Section~\ref{sec:nlo} we explain \ourMeth and how it contrasts to error-based
optimization, and provide a list of our contributions. In Section~\ref{sec:relatedwork}
we discuss related work.
 In Section~\ref{s:initialization} we describe the latent variables, and the initialization networks in Section \ref{s:init}. Section~\ref{s:expts} gives details of the experimental datasets.
Section~\ref{s:EBO} provides the details of the setup of error-based optimization, and Section \ref{s:NLO} gives the details of LiDO setup and network architecture. Finally, Section~\ref{s:eval} describes our evaluation measures, with the results presented in Section~\ref{s:results}.

\section{\ourMethodFull \label{sec:nlo}}
 
\begin{figure*}[h!t!]
\centering{
\includegraphics[width=0.999\textwidth]{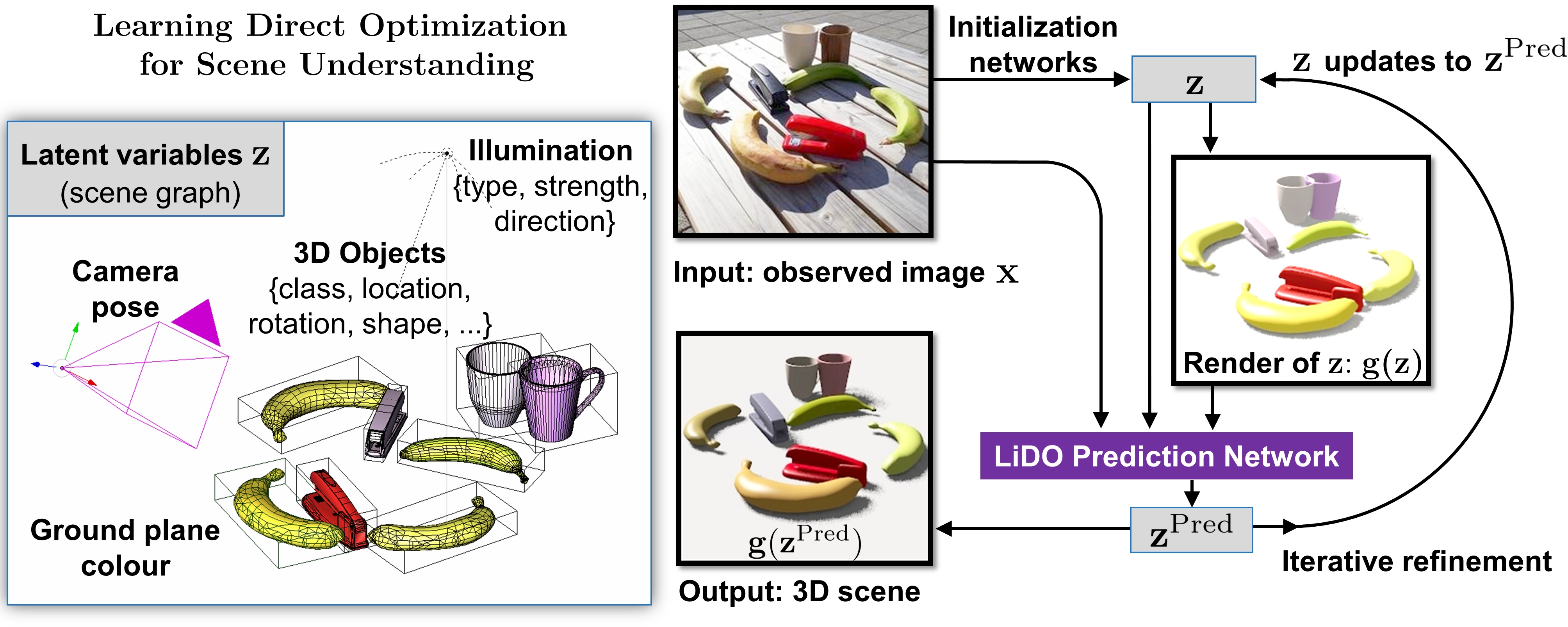}}\\
  \caption{ \ourMethodFull: given the observed image $\bfx$, the initialization of the latent variables (LVs) $\bfz$ is obtained -- then $\bfz$, the predicted image $\bfg(\bfz)$ and the observed image $\bfx$ serve as the input to the \ourMeth Prediction Network. The LVs are then updated according to the prediction and a new render is produced. The LVs are then refined iteratively driven by the Prediction Network. 
   \label{i:nlo}}
\end{figure*}

  Figure~\ref{i:nlo} gives an overview of the \ourMethodFull framework.

 In our system we use \emph{initialization networks} 
 to predict the starting configuration $\bfz_0$ based on
$\bfx$,  which also serves as the initialization for all the compared
methods.   The representation of the LVs consists of global and object
LVs and is of variable dimension, \ie $\mathbf{z} = (\VEC
z^{\mathrm{Global}}, \VEC z_{1}^{\mathrm{Object}}, \ldots, \VEC
z_{P}^{\mathrm{Object}})$, thus initialization networks can initialize
an arbitrary number of objects. The initialization
networks are described in Sections~ \ref{s:initialization} and \ref{s:init} below.

Our setup can be viewed as a kind of image autoencoder, with the
initialization networks being the encoder, the bottleneck consisting
of the LVs $\bfz$, and the graphics renderer being the decoder.
The 
optimization of $\bfz$ to improve the fit to the image is 
non-standard in autoencoders.

The LiDO method trains the \emph{Prediction Network} on data where
the current state $\bfz$ does not match the ground truth 
$\bfz_{GT}$. This was obtained in two ways: (i) from the initialization
network where, based on $\bfx$,~$\bfz_0$~and~$\bfg(\bfz_0)$, one can
predict $\bfz_{GT}$, and (ii)~by perturbing $\bfz_{GT}$ to produce
$\bfz'$, and learning to predict $\bfz_{GT}$~given~$\bfx$,~$\bfz'$~and~$\bfg(\bfz')$. 
Note that the training data requirements for Prediction Network 
are similar to that needed to train the initialization networks, and reuse the same data generator, so it has minimal marginal cost.

Given our initial prediction $\bfz_0$, a standard optimization would make steps based on some error $E$  
 where 
$E(\bfx,\bfg(\bfz_t))$ measures the
error between the image $\bfx$ and the current prediction
$\bfg(\bfz_t)$. 
To perform the optimization, one can use a gradient-based optimization (GBO).
However, due to the 
difficulties in obtaining gradients from a renderer, much work for
minimizing $E(\bfx, \bfg(\bfz))$ has used gradient-free local search
methods such a Simplex search, coordinate descent, genetic algorithms
\cite{stevens-beveridge-01}, or the COBYLA algorithm (as used in
\cite{izadinia-shan-seitz-17}).

 However, it is unknown which error measure $E$ would be most effective for simultaneously addressing issues such as misaligned objects, occlusions, textures, etc.
In contrast, \ourMethodFull procedure takes as input
$\bfx$, $\bfz_t$, and the current prediction $\bfg(\bfz_t)$, as shown
in Figure~\ref{i:nlo}, and is trained to predict
$\bfz^{\mathrm{pred}}_{t+1}$, the true latent variables corresponding
to $\bfx$, as described in Section~\ref{s:NLO}.

The key insight is that comparison of $\bfx$ and the render
$\bfg(\bfz_t)$ may yield much more information than simply the value
of the error measure $E$. 
\Red{For example, consider a scene $\bfx$ containing a mug
viewed from above. If the  prediction
of $\bfz_t$ has the overall size and position of the mug correct, but
the pose is incorrect so that the mug handle is in the wrong
place, a comparison of the two images (e.g.\ by subtraction) will show
up a characteristic pattern of differences which can lead to a large
move in $\bfz$ space.  In contrast, if the handle positions are far
apart, a small change its position will not change the error at all,
so there is zero gradient. Indeed this error will not change until there is 
an overlap between the predicted and observed handles.}

Another problem is that the optimization may be misled when the
observed image contains noisy features in a form of object textures,
shadows, etc., while LiDO Prediction Network can learn to ignore such
distractions.

Our contributions are:
\begin{enumerate}
\item   We develop a general framework for the joint refinement of all the considered LVs in a 
multi-object scene,  
including the shape and appearance of the objects,
 illumination and camera variables, 
without the need to choose a specific error metric $E$ to
  measure of the mismatch between the input image and the predicted image. 
  \item We show that \ourMeth generally produces better solutions
  in shorter time and is more stable than standard optimizers,
since  the  $\bfz$-update directly targets the optimal $\bfz$ rather than simply moving downhill.  
\item We show that \ourMeth is better able to handle mismatch between
 the data generator and the  fitted scene model, in terms of synthetic vs. real images mismatch, object shape mismatch, and mismatch due to the texture and other nuisance variables.
\end{enumerate}

 \section{Related Work}
\label{sec:relatedwork}
The task of scene understanding is a long-standing problem in  the \linebreak
computer vision literature, for which various approaches have been
developed. In recent years there has been an emphasis on methods
which make \linebreak predictions in the image frame. These include object
detection methods which output bounding-boxes  (\eg Selective Search method
\cite{uijlings2013selective}, and YOLO network \cite{Redmon_2016_CVPR}), object
instance segmentation \cite{hariharan2016object} and semantic pixel
labelling, \eg \cite{badrinarayanan2017segnet}, \cite{liu2015crf}.  An
older viewpoint in computer vision focuses on recovering 3D geometry
of the world from one of more images, such as the work of Marr
and Nishihara
\cite{marr-nishihara-78}.  Malik et al.\ \cite{malik2016three} discuss the three
``R's'' of Recognition (object detection), Reconstruction (inverse
graphics) and Reorganization (perceptual organization); they show how
these three components can interact to aid each other for scene
understanding.  Recent work on this theme includes 
pixel depth prediction \cite{li2018monocular};
exploiting 3D
geometry by placing object detections into perspective at predicted
scale and depth \cite{hoiem2008putting}; and  representing
an indoor scene by furniture items described by 3D bounding-boxes
\cite{choi2015indoor}.

Our work is carried out in the vision-as-inverse-graphics (VIG) or
analysis-by-synthesis paradigm, where the vision task is the inverse
of the rendering process. This approach not only extracts a 3D scene from the
input image, but \emph{renders} that scene to allow comparison with
the input image. VIG is a long-standing idea, see 
\cite{grenander-78} and \cite{stevens-beveridge-01},
but it can be reinvigorated using the power of deep learning for the
analysis stages. VIG can be broken down into an initialization phase,
and a subsequent refinement phase. Williams et al.\ 
\cite{williams-revow-hinton-97} is an early example of 
using neural networks for the initialization phase. 
More recent work includes \citep{tran2018nonlinear} and \cite{genova2018unsupervised} who train autoencoders for this task on face data.

The standard practice for refinement after the initialization
is to minimize some error function $E$, where the reconstruction loss
is based on a summary statistics of the image pixels. 
The error could, for example, measure the discrepancy in pixel
space, or in some other feature space like the representation obtained
in higher layers of a neural network (see e.g.\
\cite{kulkarni-whitney-kohli-tenenbaum-15}).
In contrast, \ourMeth directly predicts updates for all different
kinds of LVs together, without the need to chose a specific error
metric $E$.

Refinement within the VIG paradigm has also been considered 
by several works focusing on the specific problem of face explanation. 
This is a significantly simpler problem for optimization than 
multi-object scene explanation, as the face (a single object) 
is always present in the centre of the image and can be 
modelled by a single deformable mesh.  
 Yildirim et al.\ \cite{yildirim2015efficient} use a pixel-based error measure
when sampling parameters of a 3D face model; 
and Sch{\"o}nborn \etal \cite{schonborn2017markov} 
develop a sampling procedure over the probabilistic parameters of the 
face shape and appearance model.  Hu et al.\ \cite{hu2017efficient} split 
the problem into simpler sub-tasks and sequentially optimize 
the pose, shape, light and texture parameters. 

For multiple objects the problem is much more challenging, even in 2D,
as illustrated by \cite{jampani15_cviu} who compare various
compute-intensive MCMC methods for the relatively simple problem of 
fitting multiple colourful 2D squares. For 3D objects, most of the
methods match only the 3D geometry to the image via slow sampling procedures.  For instance Satkin \etal
\cite{satkin20153dnn} match 3D CAD models based on a set of various
similarity measures calculated using predicted surface normals,
detected edges, rendered object mask etc.; the IM2CAD method
\cite{izadinia-shan-seitz-17} aligns object shapes to the observed
image by minimizing the distance in the VGG feature space; and Zou et
al.\ \cite{zou2019complete} optimize object poses to fit to the depth
channel input by enumerating a large number of object shapes and
poses. Note these methods optimize only the geometry, and none of them
model the appearance or illumination  (for visualizations,
objects are given a colour in a post-processing step). Finally, some
works consider toy synthetic scenes, but these approaches have only
been demonstrated for scenes containing known objects with fixed sizes and
appearance (e.g.\ \cite{eslami-heess-etal-16} who consider three objects of a fixed colour, and
\cite{wu-tenenbaum-kohli-17} who consider scenes with objects from the Minecraft game), and hence have not demonstrated
applicability to real images.

\ourMeth  not only fits shapes and poses to the image, 
but makes use of a whole render of a reconstructed 3D scene, 
and refines all scene graph LVs jointly.
This idea of making use of the current render of the scene, falls into 
the area of ``auto context", which relates to 
feeding the output of a learning
machine to the input to improve results and make use of context 
information. This was studied \eg by 
Tu et al.\ \cite{tu2009auto} who used a mask of
current pixel labelling to help to improve the segmentation. 
There is very recent work by Manhardt \etal \cite{manhardt2018deep} 
and the DeepIM method \cite{li2018deepim} which describe a special 
case of \ourMeth as applied to object 
6D pose estimation (3D translation plus 3D rotation). In these
papers a neural network makes iterative updates to the pose parameters,
based on the input image and a render of the current
estimate, but only for a known, specific object of a fixed size. 
In addition, LiDO can handle 
novel instances at test time, allowing for
variable object size, shape, and texture.

\section{Latent Variables }
\label{s:initialization}

\begin{figure}[h!t!]
\centering
\includegraphics[width=0.999\textwidth, trim= 2.5cm 1cm 4.0cm 7.5cm, clip]{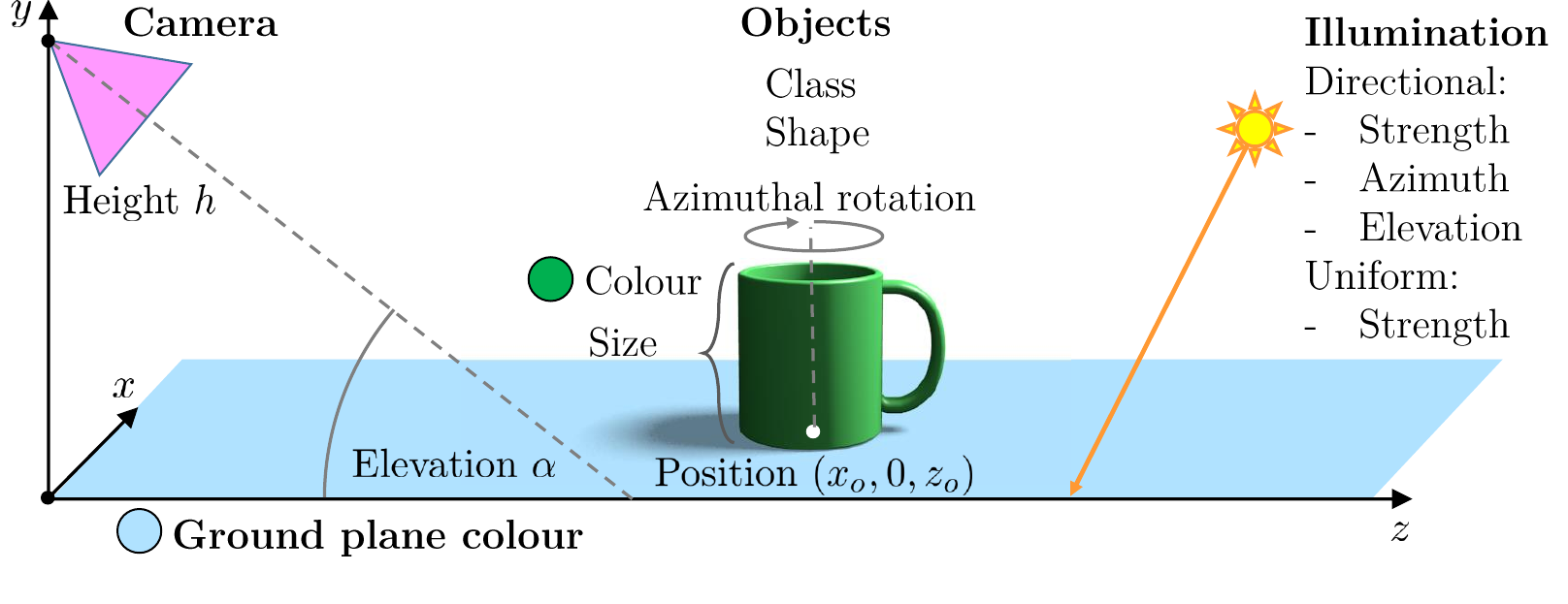}

\includegraphics[width=0.242\textwidth]{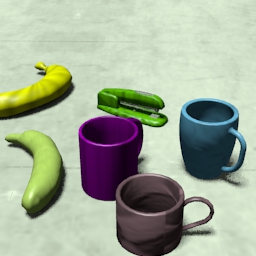}
\includegraphics[width=0.242\textwidth]{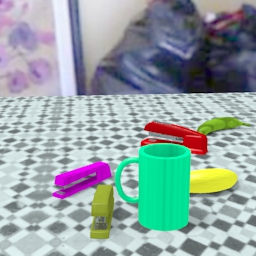}
\includegraphics[width=0.242\textwidth]{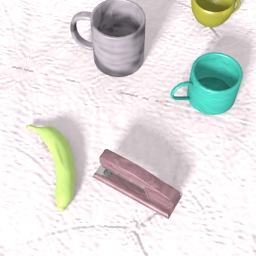}
\includegraphics[width=0.242\textwidth]{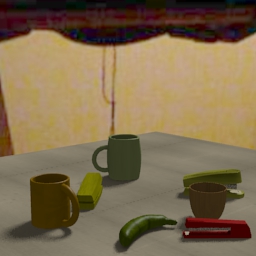}

\caption{\textbf{Top}: Diagram of the latent variables; \textbf{bottom}: examples from the Synthetic dataset, featuring a variablity in the objects present, their poses, appearance, as well as variable illumination and viewpoints. \label{i:dataset}}
\end{figure}

Our work below considers high resolution scenes  with a number of objects (from a known set
of object classes) on a \GR (table-top). See Figure~\ref{i:dataset} for the diagram of the latent variables and  example images.

\textbf{The object LVs are as follows:} for each object $o$ its associated
LVs are its class $c_o$, position \Red{$(x_o,0,z_o)$} on the \GR, size $s_o$, angle of azimuthal rotation $\phi_o$,
shape (1-of-$K$ encoding) and colour (RGB).

\textbf{The global LVs are as follows:} \GR 
RGB colour, camera LVs and illumination LVs. 
The camera is taken to be at height $y=h$ above the origin
of the $(x,z)$ plane, and to be looking at the \GR with angle
of elevation $\alpha$, with fixed camera intrinsic parameters.
The illumination
model is uniform lighting (LV: strength) plus a directional source (LVs: strength, azimuth and elevation of the source).

 The rendered objects and the
\GR have random noisy textures, and the background is a real
indoor image from NYU Depth Dataset V2\ \cite{Silberman:ECCV12}.

  The 1-out-of-K object shape encoding is a simple yet effective
shape model. As the predictions are made \emph{per detected object and per
  object class}, one could extend this to use e.g.\ shape and texture
morphable models like Blanz and Vetter \cite{blanz2003face} or later work. However,
note that the contribution of our system is demonstrating strong
performance on optimizing \emph{multiple objects} in a complex scene
(plus camera, illumination), not just one object.

\section{Initialization Networks}
\label{s:init}
Our method uses a separate network (Initialization Network) to provide a first estimate $\bfz_0$ of the values of the LVs that will be then refined.

In our framework we  define the object ``central contact point'' to be
the point located at the origin of the CAD object where it 
 is placed on the ground plane. The azimuthal rotation is specified around the vertical axis that passes though the central contact point.

The steps  
in obtaining an initial scene description  are:
\begin{enumerate}
\item Detect objects: class, central contact point
 and size; extract patches $\VEC P^{\bfx}_0 $ from input image $\bfx$ at the central contact points.
\item For each patch $\VEC P^{\bfx}_{0;p}$ with $p = 1, \ldots P$
  predict global LVs and object LVs:
  $\mathbf{z}_{0;p} = ({\VEC z_{0;p}^{\mathrm{Global}}, \VEC
    z_{0;p}^{\mathrm{Object}}})$ -- global LVs are predicted by each object. 
\item Aggregate votes for global LVs to obtain: \newline $\mathbf{z}_0 = (\VEC z_0^{\mathrm{Global}}, \VEC z_{0;1}^{\mathrm{Object}}, \ldots, \VEC z_{0;P}^{\mathrm{Object}}).$
\end{enumerate}

The above  steps make use of the detector and LV initialization networks described below, for each
patches are of size 128$\times$128 pixels. 
All the convolutional networks are trained on top of all the 13 convolutional
layers of VGG-16  network \cite{simonyan-zisserman-15}, so as to
afford transfer to work on real images.

\textbf{Object detector:}
The detector is trained to predict whether a particular object
class is present at a given location, together with object size.
Trained object detectors are  run over the input image to produce a set of detections, which are then sparsified using non-maximum suppression (NMS).

\textbf{LV initialization networks:} 
We extract an image patch
centred at each object detection, and use this to predict the ground
truth latent variables. The  networks are
applied individually to each detected object patch. All the objects
predict their own LVs as well as all global LVs. All the global LVs
are trained/predicted per object, then combined; for robustness, the
aggregation is carried out using the median function. 

\ref{a:det} gives the details of the network architecture, implementation and training.
An important aspect of the architecture is that for azimuthal rotation, we predict the rotation discretized into 18 bins of 20 degrees
(the LiDO network predicts the updates in a similar manner, discretized into smaller, 1 degree bins).
This allows multimodal predictions and thus the networks can handle the symmetries. For example for a cube object, the network should predict 4 bins every 90-degrees with similar probability, initialize at one of these, and then refine the remaining misalignment.

The outputs of the above stages are assembled into a scene 
graph. The central contact points of the detected objects  are
back-projected into a 3D scene given the predicted camera to obtain
the 3D positions. The object scaling factors are obtained from the
predicted object size and the actual distance from the camera to the object after back-projection.

\section{Stochastic Scene Generator and Experimental Datasets}
\label{s:expts}

This section present the details of the Stochastic Scene Generator (Section~\ref{s:ExSSG}),
the training and test datasets (Sections~\ref{sec:expts_trainingdata}, \ref{s:Extestdata}),
and the quality of the initialization (Section~\ref{s:ExInit}).

\subsection{Stochastic Scene Generator \label{s:ExSSG}}
For each image we sample the global LVs and the LVs of up to 7 objects
which lie on the \GR. We consider three object classes: mugs, bananas and staplers. For each object we sample its class, shape
(one-of-$K$ = 6/15/8 shapes respectively\footnote{The shapes were obtained from
ShapeNet: {\url{https://www.shapenet.org/}}; and then aligned in 3D to 
have the same position, size, and rotation.}), colour,
rotation, and scaling factor.
Since our  goal is to apply our networks to real images, the synthetic images are generated
with a rich realistic Blender\footnote{\url{https://www.blender.org/}} renderer, where we have added shadows,
realistic backgrounds and textures on the objects. The textures serve
as a noise to allow \ourMeth to work with richer
real images that may feature different kinds of surface patterns, shadows and noise.
 Textures are applied 
at a random scaling on the surface via multiplication of the initial
colour and the texture intensity. 
 Since we do not model the textures, the mean effect of
texture is absorbed into the ground truth (GT) colour. We define the GT colour to be the one that the best matches the texture, i.e.\ the mean colour of the texture. For example for the white object in black dots, the ground truth colour will be light-grey.

 The synthetic images generated by the stochastic scene
generator are rendered at  256$\times$256 pixels.  
 We sample the camera height
and elevation uniformly in the appropriate ranges: $\alpha \in
[0^\circ,75^\circ]$, $h \in [5,75]$ cm.  Illumination is represented
as uniform lighting plus a directional source, with the strength of
the uniform light $\in [0, 1]$, the strength of the directional light
$\in [0, 2]$, with azimuth $\in [0^\circ,360^\circ]$ and elevation
$\in [0^\circ,90^\circ]$ of the directional light.

To sample a scene we first select a target number of objects (between four and seven).
We then sample the camera parameters and the plane colour.
Objects are added sequentially to the scene, and a new
object is accepted if at least a half
of it is present in the image, it does not intersect other objects,
and is not occluded by more than 50\%. If it is not possible  to
place the target number of objects in the scene (\eg when a camera is
pointing downwards from a low height) we reject the scene.
For each object we sample its class (stapler, mug or banana), shape, colour,
rotation, and scaling factor so that stapler length is $\in [12,16]$cm, mug
diameter in $\in [7,10]$cm, banana length $\in [15,20]$cm.

Below we describe the process of sampling realistic colours for our
scenes. Initially we experimented with sampling from a uniform
distribution but it often results in pastel colours,  close to
gray. Therefore we use a collection of 17 predefined CSS/HTML colours
and sample a pair of them with a random mixing proportion. This samples
a variety of colours with frequent strong colours 
(where the RGB value is either close to 0 or 1), as these are common choices
for everyday objects.  Afterwards, we add a uniform noise of $\pm 0.2$ to the RGB coefficients and
clip if necessary. We use this scheme to sample colours of staplers
and mugs, for bananas we fix one of the components to be yellow, for ground plane colours we fix one of the
components to be white so as to obtain bright colours more frequently.

\subsection{Training Datasets}
\label{sec:expts_trainingdata}
 We train the initialization networks on
a dataset of 10,000 images with over 55,000 objects. To train the Prediction
Network we use data from two sources. The first (another 10,000 images
and over 55,000 objects) is obtained from the $\bfz_0$ outputs of the
initialization networks.  We paired the detected and GT objects based
on the distance of the object central contact points to the closest one of the
same class within the radius of 10\% of the image width (15\% for real
images since manual annotations are more noisy than the perfect
synthetic ones).  The second source was a dataset generated by adding a small amount of 
noise to 10,000 GT images (over 55,000 objects) to allow \ourMeth to
deal with small errors in further iterations. The noise was uniform for the continuous LVs:
$\pm$ the median error made by the initialization networks per LV. We also replaced each GT CAD shape
by a random one to train \ourMeth to work well in the case of shape-mismatch. Thus, the
\ourMeth training dataset consisted of 110,000 object examples.

\subsection{Test Datasets}
 \label{s:Extestdata}
  For all the neural networks we used
 train-validation-test splits of the synthetic dataset, using separate scenes for the Initialization Networks and LiDO Prediction Network.
Furthermore, we used a separate validation set for optimization tasks
to choose the hyper-parameters of the \ourMeth and baseline methods,
plus a final optimization test set to evaluate them
(each of 200 images, with over 1k objects).

\begin{figure}[h]
\centering

\includegraphics[width=0.2\textwidth]{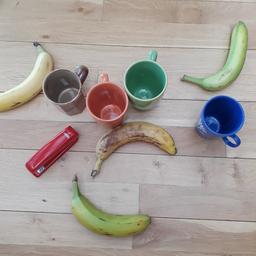}
\frame{\includegraphics[width=0.2\textwidth]{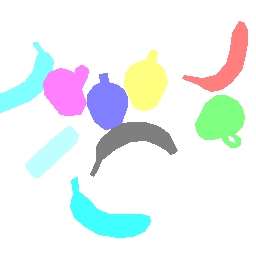}}  \quad
\includegraphics[width=0.2\textwidth]{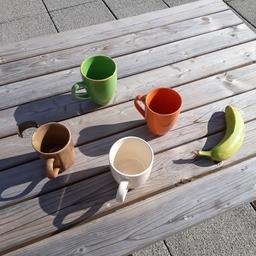}
\frame{\includegraphics[width=0.2\textwidth]{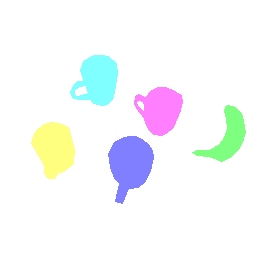}} 
  \caption{Two examples from the Real dataset (images and instance segmentation masks).}
 \label{i:datasetR}  
\end{figure}

As our aim is to understand real images, we apply the same 
methods to a dataset consisting of 135 real images with over 750 objects total. 
The manual annotations are used only for the evaluation, not for making the predictions. The real images were annotated 
with object masks and central contact points to allow
quantitative testing of the methods' performance, 
as the quantitative evaluation here can be done only in the pixel space. 
We captured the real images to feature a number of objects of the considered classes at a variety of lighting, viewpoint and object configuration conditions, see Figure~\ref{i:datasetR}.
 For each object we annotated its class, instance mask, and the central contact point using LabelMe software \cite{russell-labelme}.
 Our system renders objects on an infinite ground plane. Since the ground plane is finite for real images, we use a GT ground plane mask that is defined as the ground plane up to a
horizontal line located at the central contact point of the farthest GT
object. For real images we annotated the GT \GR mask, since sometimes the mask might not be the full plane.

\subsection{Initialization for the Test Datasets}
\label{s:ExInit} For the synthetic
dataset, objects were accurately detected
with 94.6\% precision, 94.0\% recall (94\% of objects are detected,
94.6\% of all detections are correct), and for these 57\% of the object shapes were predicted correctly.
For real images, the results
were: 97.8\% precision, 94.0\% recall, showing that the initialization networks worked similarly well for real images. All the methods
that we compare start from the same initialization with the same set
of the detected objects, and unpaired objects are treated as false
positives/negatives. The set of the instantiated objects and their shapes are kept
fixed, as these are discrete variables which are not changed during optimization with the above methods.

\section{Experimental Setup of Error-Based Optimization}
\label{s:EBO}
We compare \ourMeth to two optimization methods: gradient-based
optimization (GBO) with the best performing optimizer, and the most
effective gradient-free method which was Simplex search (denoted
\SIM). Note that these baselines were selected as the
best-performing methods out of many tested. Since these are standard search methods which require a large
number of function evaluations, we use a very fast OpenGL
renderer. \ourMeth was also configured to use OpenGL as the internal renderer during
optimization.

For error-based optimization, we compute the
match between the actual and rendered image pixels (RGB intensities being between 0 and 1) using a robustified
Gaussian likelihood model with standard deviation $\sigma=0.1$ and inlier probability $\alpha = 0.8$, as in \cite[eq.\ 3]{moreno-williams-nash-kohli-16}. 
 The observed image $\mathbf I^O$ and the rendered image $\mathbf I^R(\mathbf x)$ have $P$ pixels and are represented as a vector of a length $3P$, with $P$ values for each RGB colour channel. Then, for each pixel-channel $i$, $p( \mathbf I^O_i|\mathbf z)$ is given as in \cite{moreno-williams-nash-kohli-16} by:
\vspace{-6mm} 
 
\begin{align}
p( \mathbf I^O_i|\mathbf z) = \alpha \mathcal{N}( {\mathbf I}^O_i; {\mathbf I}^R_i(\mathbf z), \sigma^2) + (1-\alpha)\mathcal{U}( \mathbf I^O_i).
\end{align}

For GBO we use a differentiable OpenGL
renderer\footnote{\url{https://github.com/polmorenoc/inversegraphics}}
based on OpenDR: Differentiable Renderer \cite{loper2014opendr},
extended to simplify rendering multiple objects. The
approximate derivatives of the likelihood computed by OpenDR are fed to an
optimizer.
 To facilitate refinement we use anti-aliasing with 8
samples per pixel to make the gradients more accurate and the
likelihood function smoother. 

We performed experiments with several optimizers and most of them
converged poorly (\eg L-BFGS-B). We found Truncated Newton Conjugate-Gradient (TNC) to
considerably outperform other gradient-based methods, with the Nonlinear Conjugate
Gradient
optimizer\footnote{\url{http://learning.eng.cam.ac.uk/carl/code/minimize}}
being the only other one that usually converged well (yet worse than TNC, so we use TNC for GBO).

For gradient-free methods, we found the Simplex (Nelder-Mead) optimizer worked well and significantly better than COBYLA, Simplex also often performed better and faster than GBO.

Setting proper bounds is crucial for proper optimization as the
stepsize is scaled by the distance between LV boundaries, hence for
all the LVs we set the bounds to the respective ranges as used in the
scene generator.

Following the work of \cite{moreno-williams-nash-kohli-16}, we fit
subsets of the search variables sequentially.  The LVs are fit in the
following order: ground plane colour, object colours (each object
separately), object poses (each object separately), illumination and
the camera.  We experimented with fitting each object's LVs together, all the object LVs together, and also all LVs together, but it worked a lot less well and overall slower, because the number of variables is larger and likely the
optimization landscape is thus more complex.

\section{Experimental Setup of \ourMethodFull \label{s:NLO}}

We ran the initialization networks on the synthetic dataset and took object
detections and the associated errors as the new dataset for training
 \ourMeth. For an image $\bfx$ with ground truth $\bfz_{GT}$
we obtained a set of image patches $\VEC P^{\bfx}_{0}$ extracted at the object
detections and the corresponding rendered patches $\VEC P^R_{0}$ from
the render $\bfg(\bfz_0)$. From each pair of patches 
$\VEC P^{\bfx}_{0;p}$ and $\VEC P^R_{0;p}$, $p = 1, \ldots, P$ and the corresponding
$\bfz_0$ the 
\emph{Prediction  Network} CNN is trained to predict the object-specific 
GT variables $\bfz^{\mathrm{Object}}_{GT;p}$ and the global
GT variables  $\bfz^{\mathrm{Global}}_{GT}$. 
Example errors are that an object could be larger,
  the camera located higher.

The Prediction Network takes as input both the observed and the
rendered image  patches (128$\times$128 pixels, down-sampled to 64 by 64 resolution), plus their difference. The RGB channels are
stacked together giving an input size of $64 \times 64 \times 9$. This input is
followed by a number of convolutional layers shared across all the
LVs. Shared layers are followed by LV-set specific convolutional layers, and finally a few
 fully-connected layers, each concatenated with the
current estimate of $\bfz_p$.
For each object patch, the current estimate of $\bfz_p$ (standardized) input consists of: object LVs (discrete class and shape one-hot-encoded), global LVs (as predicted by the object, denoted $G_O$), global LVs (as used in the render after voting of all the objects, denoted $G_V$), plus their difference $G_O - G_V$. 
The whole network for all the LVs is trained together.

To allow convergence, the object pose is trained in the object's current
coordinates/frame (to predict the change in object position and rotation). CNNs are trained together with an L1+L2 error for
continuous LVs and categorical cross-entropy   for discrete ones. We add
the L1 loss to the L2 loss, as when making predictions multiple times
small errors aggregate, and L1 appropriately punishes small errors
during training. To calculate the overall network loss we sum up each L1+L2 loss and the cross-entropy loss (for which we used a scaling factor 0.2). We use the Adam  optimizer with learning rate 0.0003.

 We trained the Prediction Network for 20 epochs, this took 15 minutes/epoch on a single GPU. Note that we reuse the same scene generator, and that training time is no more than the training time of the initialization networks. 

Afterwards, we run the refinement for $T$ iterations as
summarized below:
 
\noindent \texttt{\textbf {for}} $ t \in 0$\textbf{..}$(T-1)$\textbf{:}
\begin{enumerate}
\item  Take patches $\VEC P^{\bfx}_t, \VEC P^R_t$ from the input image
  $\bfx$ and the render $\bfg(\mathbf{z}_t)$ 
\item Predict $\mathbf{z}_{t+1;p}^\mathrm{pred}$ given each $\VEC
  P^{\bfx}_{t;p}, \VEC P^R_{t;p}, \mathbf{z}_{t;p}$ using the Prediction Network \\ (clip if outside of the range, e.g. colour not in [0,1]).
\item Update $\bfz_{t+1;p}  = \bfz_{t;p} + \mu_t
  (\bfz^{\mathrm{pred}}_{t+1;p} - \bfz_{t;p})$.
\item Aggregate global LVs to produce $\VEC z_{t+1}$ (in the same manner as in Section~\ref{s:init}).
\end{enumerate}

Setting the step size $\mu_t = 1$ would move from $\bfz_t$ to
    $\bfz^{\mathrm{pred}}_{t+1}$, but we have found that in the case
        of multiple updates, using a $\mu_t < 1$ which decreases with $t$
          leads to better performance than keeping $\mu_t$ fixed. We set $\mu_t = 1/(t+a)$.  The hyper-parameter  $a=2$ was selected using the validation set split. In the experiments below we run the \ourMeth iteration for a fixed number of $T = 30$ steps to show the convergence curve, but it would be easy to use a termination condition
$|\bfz_{t+1} - \bfz_{t}| < \epsilon$.

\begin{table*}[h!t!]
\centering
\begin{tabular}{ c c c c   p{1.2cm}| c }\toprule
\multicolumn{6}{c}{\bf LiDO Prediction Network} \\

 Position & Size & Azimuth & Lighting & Camera & Colour (Ob/Gr)  \\
\cmidrule{1-5}\cmidrule{6-6}
\multicolumn{6}{c}{Input $64 \times 64 \times 9$ (2 images plus their difference, stacked)}\\
\cmidrule{1-6}
\multicolumn{5}{c|}{ C-32-3  \emph{(shared)}}  & C-32-3 (stride 2)\\
\multicolumn{5}{c|}{C*-64-3 \emph{(shared)}} & C-64-3 (stride 2)\\
\multicolumn{5}{c|}{MaxPool-2 \emph{(shared)}} \\
\multicolumn{5}{c|}{C*-128-3 \emph{(shared)}} & C*-128-3 (stride 2) \\
\multicolumn{5}{c|}{MaxPool-2 \emph{(shared)}} \\
\cmidrule{1-5}\cmidrule{6-6}

\multicolumn{5}{c|}{C*-64-3 \emph{(separate per network)}}& \\
\multicolumn{5}{c|}{MaxPool-2 \emph{(separate per network)}}& \\
\multicolumn{5}{c|}{C-32-3 \emph{(separate per network)}}& \\
\multicolumn{5}{c|}{3 $\times$ Fz-40 \emph{(separate per network)}} & 3 $\times$ Fz-40\\

Fz-2 & Fz-1 &  Softm.(Fz-360) & Fz-5 & Fz-2 & Fz-3\\
\bottomrule
\end{tabular}
\caption{The configurations of the LiDO Prediction Network, the whole network is trained together. Layer description (where N denotes the number of units and K the filter size), is as follows: 
1) \emph{Convolutional layer}: C-N-K;
2)  \emph{Fully connected layer: F-N};
3)  \emph{Fully connected layer, with its input concatenated with the current LVs $\mathbf{z}$}: Fz-N;
4) \emph{Max-pooling layer}: MaxPool-K;
5) \emph{Softmax output layer on top of a linear layer X}: Softm.(X).
 Non-colour networks: the first 5 layers are shared across all the 5 sub-networks, all the sub-networks on top of them have the same setup. Colour networks are simpler and have fewer layers, and use \emph{ReLU} activations, while non-colour networks use \emph{tanh} activations. For lighting we predict: (uniform component strength, directional component: strength, elevation, sin(azimuth), cos(azimuth)). We use dropout with $p = 0.5 $ after the  convolutional layers denoted with *.
}
\label{t:layers1}
\end{table*}

Table~\ref{t:layers1} shows the network configurations used for \ourMeth. The implementation is in Python (TensorFlow). Again, the region outside the image frame is given as value 0. The [0,255] image dataset values had their mean subtracted and were divided by 100.

When producing the OpenGL renders for the \ourMeth prediction, 
we needed to take care because \ourMeth has been trained to 
predict $\bfz$'s that specify a scene for the Blender renderer.
While the geometry LVs (objects present, their shape/poses, camera
etc.) are common for both the renderers, the optimal colours in OpenGL
differ in brightness to Blender, since the OpenGL renderer cannot produce
shadows, and has to explain shadowing (\eg that mugs are dark inside when the light comes
from the side) with a lower colour brightness. 
To render an OpenGL image for the \ourMeth prediction with Blender
colour LVs, we adjust the brightness colour of each object and the
ground plane. We do so by scaling the RGB colour by the ratio of the
means of brightness calculated at the pixels of the predicted object
mask, for \ourMeth's OpenGL render and the observed image. We can do
this as it uses only the \emph{predicted} masks, \eg this would be
equivalent to the final iteration of minimizing 
the MSE w.r.t.\ the colour LVs.

\section{Experimental Evaluation Measures}
\label{s:eval}

 This section gives the details of the evaluation measures of our interest, 
 in the latent space (Section~\ref{sec:evaluationLVs})
   and in the image space (Section~\ref{sec:evaluationImage}).
   
\subsection{Evaluation of the LVs} 
\label{sec:evaluationLVs}
For the synthetic dataset, 
we evaluate the improvement in the LVs for
all the methods. We consider suitable evaluation measures specific 
for the seven different LV-sets, as outlined below.

Object LVs: Object position error is a distance between object central contact
points in the image.
Object size is the size of the projected object in the image frame,
the error is the relative size difference.
For azimuthal rotation we measure the absolute
angular difference between the prediction and ground truth, but with
wrap-around, so the maximum error is $180^\circ$.  
 The error metric of the object
colour  is the RMSE of normalized RGB components (computed as $R/(R+G+B)$ etc.).

Global LVs: Ground plane colour is evaluated as for object colour above. The
lighting is projected onto a sphere and evaluated at 313 
points uniformly-distributed on the sphere, then normalized; the
error is RMSE. To assess the camera error, we place a (virtual) checkerboard in the
scene, and compute the RMSE of the errors between the GT and 
predicted positions of the grid points in the image, as used
by \cite{romaszko-williams-moreno-kohli-17}.

The multiplicative interaction between illumination
and colour introduces a problem when evaluating them separately; by
using the normalized metrics above we overcome this issue. The joint
result of both factors is directly available via pixel intensities (and is compared via MSE).

\subsection{Evaluation in the Image Space (2D Projection, Pixels)}
\label{sec:evaluationImage}
We compare the observed and predicted images using the Intersection-over-Union (IoU)
 of the predicted and GT masks (of objects and ground plane), and MSE of pixel
intensities calculated at the GT mask (of objects and ground plane).
We can evaluate these measures for both synthetic and real datasets.
Note that the IoU of the ground plane assesses differences in the
present, missing and superfluous objects.  The background (the 
part of the image not
belonging to the \GR or the object masks) is excluded from the explained
pixels. Note that each MSE is calculated at the same pixels for all the methods,
as these are calculated only at the GT masks.

\section{Results \label{s:results}}

 This section provides the results:
 Section~\ref{s:ResEvLVs} presents the evaluation in terms of the latent variables and shows illustrative examples for the synthetic dataset; 
 Section~\ref{s:ResEvImage} presents the evaluation in the image space for both the synthetic and real datasets,
 and shows illustrative examples for the real dataset.

\begin{figure*}[h!t!]
\centering
\edef\trimoptions2{trim = 0.6cm 0.8cm 0.8cm 0.1cm, clip,width=0.465\textwidth}
\edef\trimoption{trim = 0.4cm 0.8cm 0.8cm 0.1cm, clip,width=0.465\textwidth}
\begin{tabular}{l}
\expandafter\includegraphics\expandafter[\trimoption]{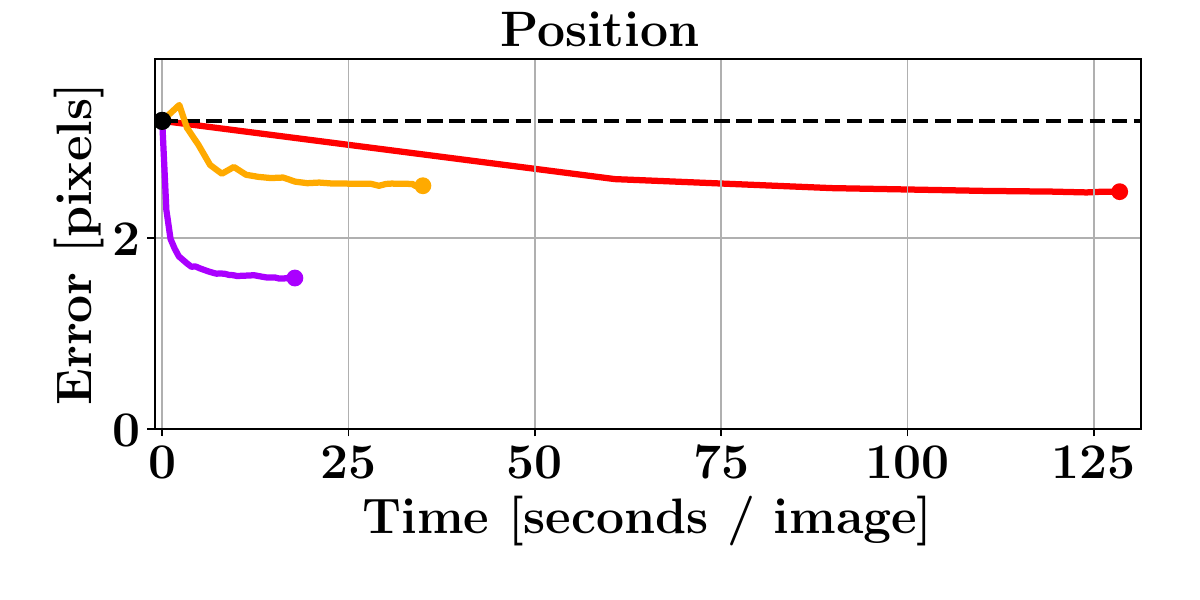}\hspace{0.5cm}
\expandafter\includegraphics\expandafter[\trimoption]{"img/plots_v2/TESTPLOT_Objectcolo"}\\

\expandafter\includegraphics\expandafter[\trimoption]{"img/plots_v2/TESTPLOT_Objectsize2"}\hspace{0.5cm}
\expandafter\includegraphics\expandafter[\trimoption]{"img/plots_v2/TESTPLOT_Azimuthalr"}\\

\expandafter\includegraphics\expandafter[\trimoptions2]{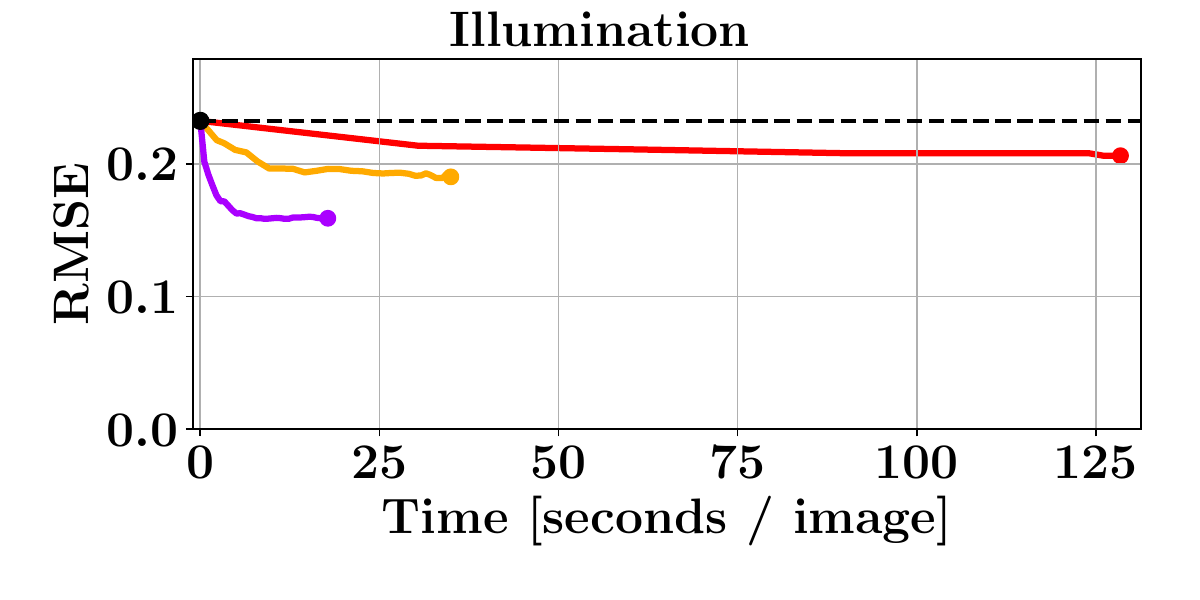} \hspace{0.4cm}
\expandafter\includegraphics\expandafter[\trimoptions2]{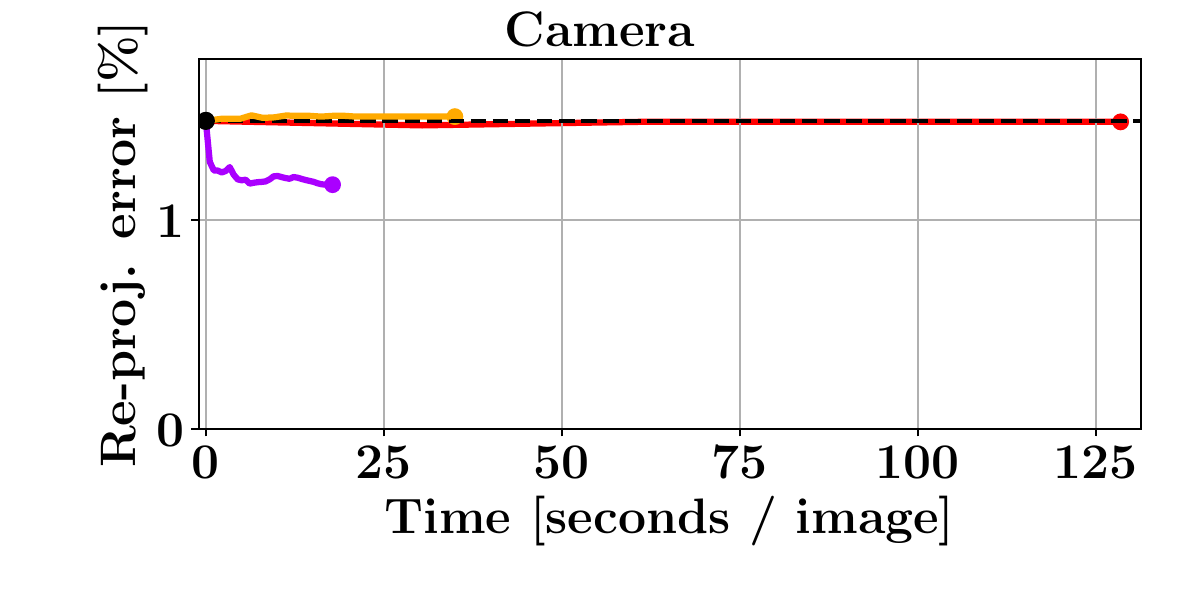}\\

\expandafter\includegraphics\expandafter[\trimoptions2]{"img/plots_v2/TESTPLOT_Groundplan"}\hspace{1.5cm}
\hspace{0.5cm}\raisebox{0.2cm}{ 
\includegraphics[width=0.21\textwidth,clip,trim=13cm 3cm 1.2cm 4cm ]{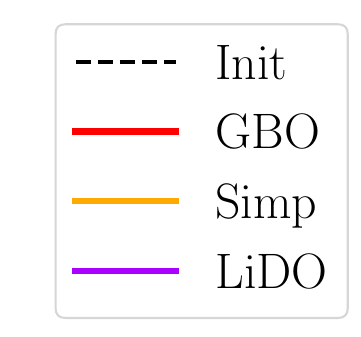}}
 \end{tabular}

  \caption{  Median errors vs time in seconds (top: object LVs, bottom: global LVs). All three methods (GBO, \SIM, \ourMeth) start from the initialization error (Init) located at the black dashed line. Note the rapid convergence of  \ourMeth.
     \label{i:res2}}
\end{figure*}

We ran the methods long enough to allow convergence: 50 iterations for GBO, 100 iterations for Simplex, and 30 for \ourMeth,
see Figure~\ref{i:res2}. 
All the times shown are for a 4-core CPU for all the methods, to make the comparison fair  \ourMeth
 is also executed on a CPU, including the
 CNNs\footnote{Although the CNNs run on GPUs were a few times faster,
   this did not affect the overall speed significantly since rendering and other modules take most of the time.}.

\begin{table}[t!h!]
\setlength{\tabcolsep}{4pt}
\centering
\begin{tabular}{c}
  \begin{tabular}{l|lc|ccl|ccl}

   & \multicolumn{2}{c|}{Initialization} &   \multicolumn{3}{c|}{Improvement[\%]} & \multicolumn{3}{c}{Impr.\ hard cases[\%]}\\
           \midrule
    LVs name  & Err. & Unit  & GBO   & Simp  & \ourMeth   & GBO   & Simp  & \ourMeth \\
        \midrule
        Object position & 3.23 & pix &   22.9  & 21.0  &  \bf 51.0*  &   76  & 71  &  \bf 92  \\ 
Object colour & 0.027 & RMSE &  14.0  & 12.4  &  \bf 59.1*  &    70  & 67  &  \bf 95  \\ 
Object size & 7.08 & \% &  \phantom{0}9.7  & 32.2  &  \bf 48.7*  &   72  & 70  &  \bf 87  \\ 
Object azimuth & 9.15 & deg. &   20.3  & 32.6  &  \bf 33.1  &   62  & 62  &  \bf 67  \\  
Illumination & 0.23& RMSE &  11.3  & 18.1  &  \bf 31.6* &   75  & 80  &  \bf 86  \\  
Camera & 1.47 & r.e. &   \phantom{0}0.3  & \phantom{0}-1.3\phantom{.}  &  \bf 20.7*  & 37  & 45  &  \bf 80  \\  
Ground colour  & 0.021 & RMSE &  68.2  & 42.4  &  \bf 73.6   &  80  & 66  &  \bf 92  \\
    \bottomrule
    \noalign{\vskip 0.5mm} 
  \end{tabular}
\end{tabular}
\caption{Results of the Initialization (left), ``Improvement''
  (centre) and ``Improved hard cases'' (right).
  \textbf{Initialization:}  median errors as per the evaluation
  measures given in Section~\ref{sec:evaluationLVs}, units are:
  pix -- pixels, deg.\ -- degrees,
r.e -- re-projection error; \newline  \textbf{Improvement}: each value indicates how much (in \%) the median error across observations was lower after the refinement compared to the median error of the initialization. Ground truth would give 100\% improvement,  {\bf{*}} denotes a statistically significantly better method.  
\newline \textbf{Improved hard cases}: for ``hard cases'' we considered the worst 50\% of the  initializations, individually per each LV-set. We show percentage of the hard cases that have been improved (see also plots in Figure \ref{i:BAs}). 
\label{t:res2}}\end{table} 

\subsection{Results: Evaluation of the LVs on the Synthetic Dataset}
\label{s:ResEvLVs}

Table~\ref{t:res2} (centre) shows the percentage improvement of the median error of each of the methods (GBO,
\SIM, \ourMeth) over the median error of the initialization (Init, left). For all  seven evaluation measures \ourMeth  outperforms both GBO and
\SIM.  For four out of seven LV sets the \ourMeth improvements are at
least two times higher than the competitors (on illumination, camera,
object: colour, position). We calculate all the different metrics as in
Section~\ref{sec:evaluationLVs},\ e.g.\ deviation in pixels for object position or angle
in degrees for rotation.  However, since
all the LVs are in different units, we compare the percentage
improvement over the initialization.

\begin{figure*}[ht]
\edef\www{width=0.16\textwidth}
\renewcommand\arraystretch{0.5}
\begin{tabular}{@{}c@{\hspace{1mm}}c@{\hspace{2mm}}c@{\hspace{1mm}}c@{\hspace{1mm}}c@{\hspace{1mm}}c@{}}

\bf Observed  & \bf Initializat.  &\bf GBO & \bf Simplex & \bf LiDO & \bf LiDO-R  \\

  \expandafter\includegraphics\expandafter[\www]{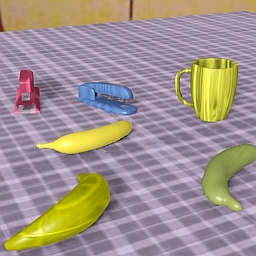} & 
  \begin{overpic}[width=0.16\textwidth]{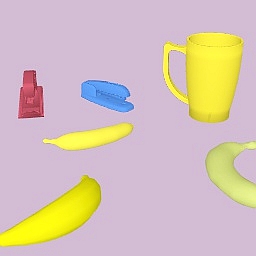}
     \put(0,0){\includegraphics[width=0.16\textwidth]{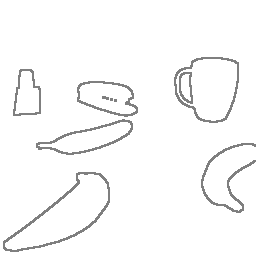}}
  \end{overpic}&

  \begin{overpic}[width=0.16\textwidth]{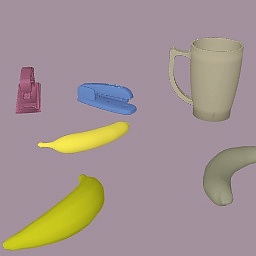}
     \put(0,0){\includegraphics[width=0.16\textwidth]{img/imgs/220CO}}
  \end{overpic}&
    \begin{overpic}[width=0.16\textwidth]{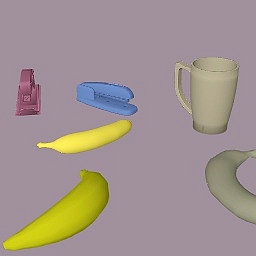}
     \put(0,0){\includegraphics[width=0.16\textwidth]{img/imgs/220CO}}
  \end{overpic}&
  \begin{overpic}[width=0.16\textwidth]{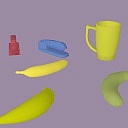}
     \put(0,0){\includegraphics[width=0.16\textwidth]{img/imgs/220CO}}
  \end{overpic}&
    \begin{overpic}[width=0.16\textwidth]{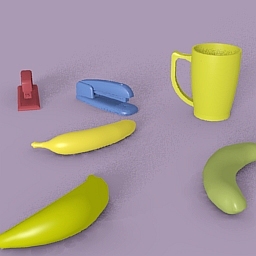}
  \end{overpic}\\

  \multicolumn{6}{p{0.98\textwidth}}{ \footnotesize{ 
{ 
 \bf Good initialization}: given a good initialization all the methods
usually converge well, \eg 4 leftmost objects, for the two rightmost
objects (mug and banana) where the initialized masks are less
accurate, only \ourMeth fits the colours properly.
}}\\

\\

\expandafter\includegraphics\expandafter[\www]{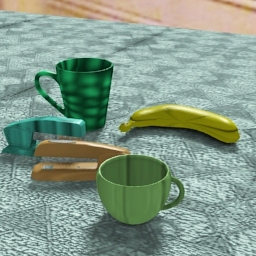} & 
  \begin{overpic}[width=0.16\textwidth]{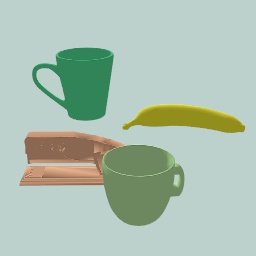}
     \put(0,0){\includegraphics[width=0.16\textwidth]{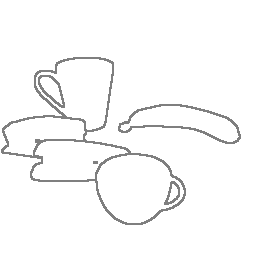}}
  \end{overpic}&

  \begin{overpic}[width=0.16\textwidth]{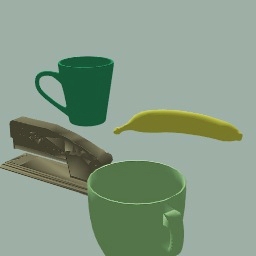}
     \put(0,0){\includegraphics[width=0.16\textwidth]{img/imgs/377CO}}
  \end{overpic}&
    \begin{overpic}[width=0.16\textwidth]{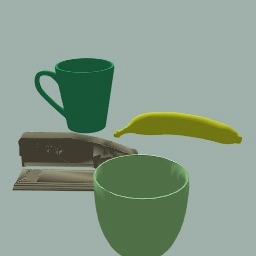}
     \put(0,0){\includegraphics[width=0.16\textwidth]{img/imgs/377CO}}
  \end{overpic}&
  \begin{overpic}[width=0.16\textwidth]{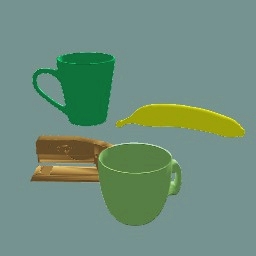}
     \put(0,0){\includegraphics[width=0.16\textwidth]{img/imgs/377CO}}
  \end{overpic}&
    \begin{overpic}[width=0.16\textwidth]{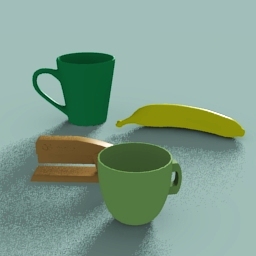}
  \end{overpic}
  \\
\multicolumn{6}{p{0.98\textwidth}}{ \footnotesize{  {\bf{Textures and shadows}}:  There are two staplers in the
input image and the blue one was not detected as it is hardly visible.
For GBO and Simp the pink stapler converges wrongly, and the same happens for
the front mug, which enlarges to explain the shadow. \ourMeth is robust to
such distractors, note for these objects the CAD shapes are different than observed.}} 
\end{tabular}
  \caption{Example runs for Synthetic dataset, showing from left: the observed input image, the
    initialization (OpenGL), and images after refinement for GBO, Simplex and LiDO using OpenGL renderers, and LiDO using Blender renderer (LiDO-R). We overlay black contours of the ground truth object masks on top of each OpenGL image to ease the comparison.
    \label{i:examp}}
\end{figure*}

To assess the statistical significance we conduct a paired test on the
errors derived from each image (for global LVs) or object (for object
LVs), using the Wilcoxon signed-rank test, at the significance level
0.05. For these \ourMeth outperforms GBO for 6 out of 7 
LVs, and \SIM also for 6 out of 7 LVs. 
This is because GBO does well with
ground-plane colour since the objective it minimizes are the
differences in pixel intensities between the input image $\bfx$ and
the render $\bfg(\bfx)$, and \SIM performs  similarly to \ourMeth for the
azimuthal rotation, but much worse for all other LVs.

Figure~\ref{i:res2} shows the evolution of the median errors over time 
for the seven error measures; it is notable that  \ourMeth  
 obtains a lower error in much shorter time; for error-based methods 
 since we do iterations sequentially for each object, we report the time for
each iteration as the average time of reaching it.

The main reason why LiDO is faster is that it requires only a single
render of the scene per update of all the LVs, and all the configurations are accepted. 
In contrast standard optimizers accept only a fraction of the rendered 
configurations, as these search over the error landscape 
and usually need several renders (e.g.\ within a line search for 
GBO to choose the step-size) before accepting a new configuration.
For our datasets, Simp accepts only approximately 30\% of the
rendered configurations it investigates, and GBO 7\%. Recall also that Simp and GBO
update only subsets of $\bfz$ per search step, as we found (see Section~\ref{s:EBO})
that fitting all LVs together worked less well and was slower overall.

\begin{figure*}[h!t!]
\centering
\edef\trimoptions2{trim = 0cm 0cm 0cm 0cm, clip,width=0.495\textwidth}
\edef\trimoption{trim = 0cm 0cm 0cm 0cm, clip,width=0.495\textwidth}

\begin{tabular}{l}
\expandafter\includegraphics\expandafter[\trimoption]{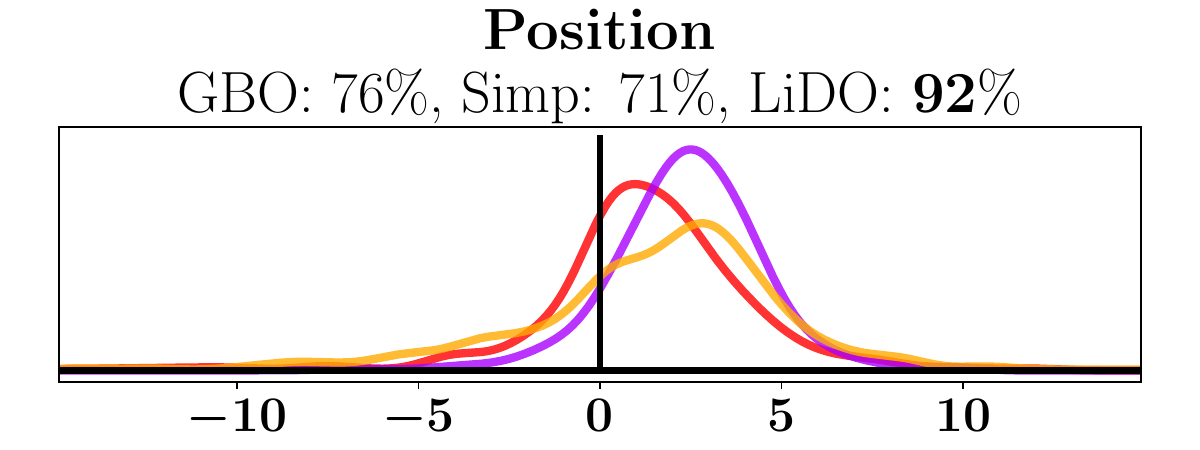}
\expandafter\includegraphics\expandafter[\trimoption]{"img/BAs/TESTHIST_Objectcolo"}
\\
\expandafter\includegraphics\expandafter[\trimoption]{"img/BAs/TESTHIST_Objectsize2"}
\expandafter\includegraphics\expandafter[\trimoption]{"img/BAs/TESTHIST_Azimuthalr"}\\

\expandafter\includegraphics\expandafter[\trimoptions2]{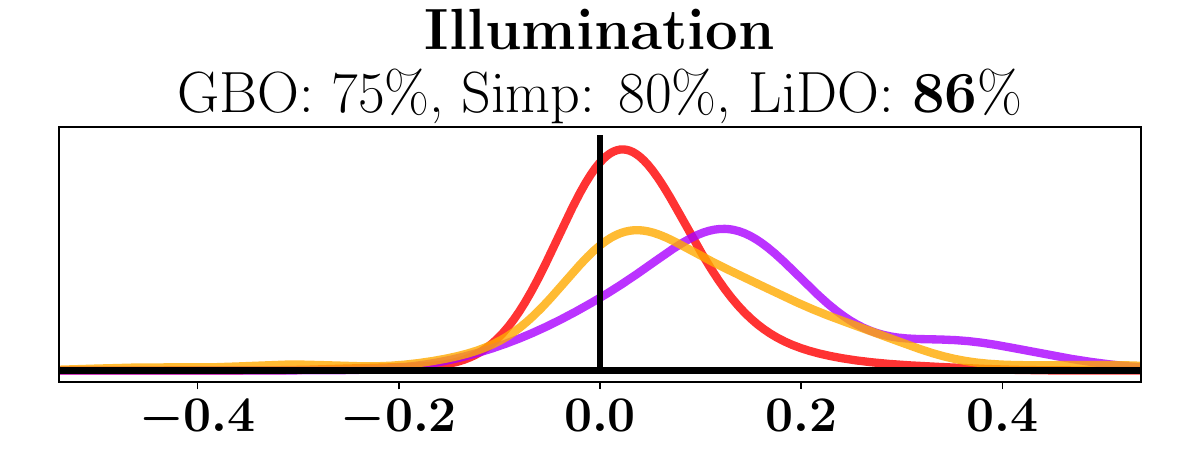} 
\expandafter\includegraphics\expandafter[\trimoptions2]{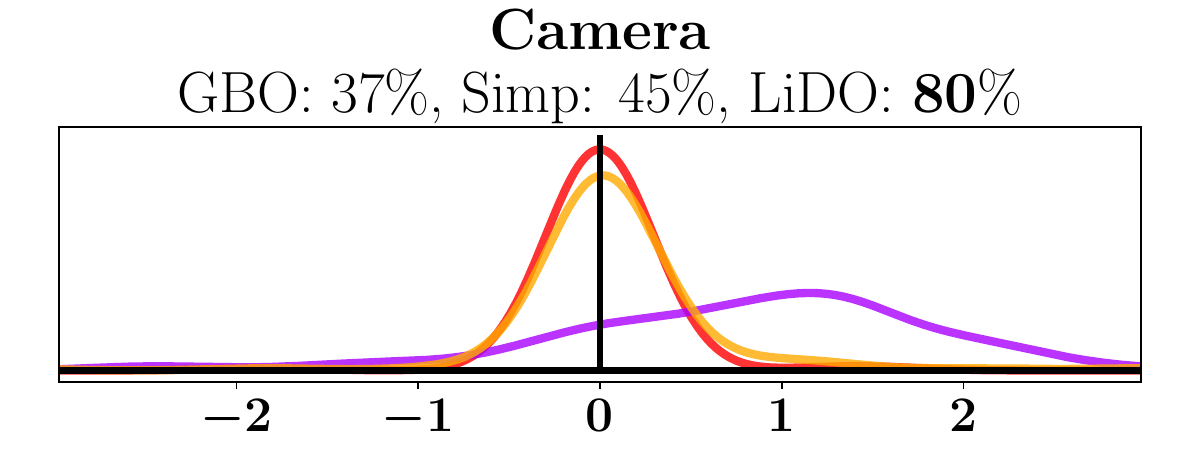}\\

\expandafter\includegraphics\expandafter[\trimoptions2]{"img/BAs/TESTHIST_Groundplan"}\hspace{1.0cm}
\hspace{0.8cm}\raisebox{0.25cm}{\includegraphics[width=0.19\textwidth,clip,trim=14cm 3.5cm 1.5cm 5.5cm ]{img/plots_v2/legend}}
 \end{tabular}

\caption{The distribution of the differences of the initial and final errors ($\mathbf{e}_{init} - \mathbf{e}_{final}$) of the methods for each of the seven LV-sets for hard cases -- kernel density estimate. Improved cases are on the right of the 0-line, the less mass on the left the better. For all the LV-sets LiDO performs  better than the baselines, note LiDO usually has much less mass on the left side of the plot than the baselines.}
\label{i:BAs}
\end{figure*}

Figure~\ref{i:examp} shows example runs, showing both success and
failure cases, see textual descriptions under each image set. 

More examples of the fitting are given
 in \ref{a:moresyn}, and the video of the fitting 
at: \url{https://youtu.be/Axc0G8IggVU}.

For evaluating the robustness and convergence of the optimizers, the most interesting cases are the ones for which the  initialization is not accurate. We performed  an additional study to measure the performance for difficult cases. For such ``hard cases'' we considered the worst 50\% of the  initializations, individually per each LV-set, these ranged from average to poor initializations. Since we have 200 test images, for each LV-set we used the worst 50\% of initializations so as to keep the sample sizes sufficiently large, this led to 532 object LVs and 100 global LVs. 

We calculate the percentage of the hard cases that are improved, these are shown in Table \ref{t:res2} (right). LiDO performs well and much better than the other methods; for 3 out of 7 LV-sets more than 90\% cases are improved. For GBO and Simp none of the individual results exceeds 80\%. 

We can get a more fine-grained view of the performance by
considering the distribution of the changes of the errors, 
see Figure~\ref{i:BAs}. For each LV-set and for each method 
we show the kernel density estimate (KDE)\footnote{for KDE kernel, we used Gaussian with a standard 
deviation of $\mathbf{\Delta}_{\texttt{max}}/20$ for object LVs 
and $\mathbf{\Delta}_{\texttt{max}}/10$ for global LVs -- the different 
values due to the different sample sizes.} of the distribution of the differences $\mathbf{\Delta} = \mathbf{e}_{init} - \mathbf{e}_{final}$
of the initial and final errors. Values on the right of the 
vertical line indicate an improvement. For 6 out of 7 
LV-sets, LiDO has a much lower amount of mass  on the 
left-hand-side than the baselines. For example, for the Position LVs it is only 
8\% ($100\% - 92\%$), while these are 24\% and 29\% for GBO and Simp respectively. 
For the Camera LVs only LiDO makes noticeable improvements. For Azimuthal rotation the 
performance of all the methods is similar.

\subsection{Results: Image-Space Evaluation for the Synthetic and Real Datasets}
\label{s:ResEvImage}
Results for the Synthetic dataset for image-space measures are given in
Table~\ref{t:resSPIX}. To allow an equal comparison of three 
optimizers, all three methods use the OpenGL renderer. For all the measures \ourMeth outperforms the
other methods, and particularly \ourMeth works much better for IoU
measures. Note that due to unrealisable textures and shadows, the minimal (OpenGL GT) MSE errors are above 0, these are 11.7 for objects, and 8.4 for the ground plane.

\begin{table}[h!] 
\centering
\begin{tabular}{lccccl}

 \toprule
Measure  & INIT  & GBO  & \SIM  &  \ourMeth    \\ 
        \midrule
IoU [ob]  &  66.2  &  73.1  &  74.9  &  \bf  78.9*  & \multirow{2}{*}{$\uparrow$ } \\ 
  IoU [gr]  &  86.4  &  88.7  &  88.8  &  \bf  91.3*  &  \\ 
  \midrule
MSE [ob]  &  54.1  &  29.2  &  26.4  &  \bf  21.8*  & \multirow{2}{*}{$\downarrow$ } \\ 
  MSE [gr]  &  34.1  &  12.8  &  13.6  &  \bf  11.9\phantom{0}  &  \\ 
  \bottomrule
\noalign{\vskip 1mm} 
\end{tabular} 
\caption{Results of the pixel evaluation for synthetic dataset. Mean
  IoU  (in \%) and MSE $(\times 10^3)$ for the objects [ob] and ground plane [gr]. The arrows indicate whether higher or lower values are better.  {\bf{*}} denotes a statistically significantly better method, using the
Wilcoxon signed-rank test, at the significance level 0.05.
}
\label{t:resSPIX}
\end{table}

\begin{table}[h!]
\centering
 \vspace{5mm} 
\begin{tabular}{lccccl}
 \toprule
Measure  & INIT  & GBO  & \SIM  & \ourMeth  & \\ 
        \midrule
IoU [ob]  &  60.9  &  63.2  &  61.5  &  \bf  71.4*  & \multirow{2}{*}{$\uparrow$ } \\ 

  IoU [gr]  &  87.6  &  87.3  &  86.1  &  \bf  91.0*  &  \\ 
  \midrule 
MSE [ob]  &  79.1  &  42.6  &  46.2  &  \bf 35.9*   & \multirow{2}{*}{$\downarrow$ } \\ 

  MSE [gr]  &  69.1  &  27.5  &  30.5  &  \bf  19.2*    &  \\ 
  \bottomrule 
  \noalign{\vskip 1mm} 
  \end{tabular}
\caption{Results of the pixel evaluation for real dataset. Mean IoU (in \%) and MSE $(\times 10^3)$ for the objects [ob] and \GR [gr]. {\bf{*}} denotes a statistically significantly better method.
}
\label{t:resRPIX}
\end{table}

Results for Real Dataset are given in Table~\ref{t:resRPIX}. Since
real images are more noisy and difficult, GBO and \SIM work poorly for IoU
(there is a very minor improvement for objects, and no improvement for
the ground plane).  All the methods improve the pixel colours (MSE),
but note this is because the pixel match is an explicit error measure
for GBO and Simp.  \ourMeth, which has been trained on
synthetic data, 
transfers to work better with real images for all four measures.

\begin{figure*}[h!t!]
\begin{center}

\edef\www{width=0.16\textwidth}
\renewcommand\arraystretch{0.5}
\begin{tabular}{@{}c@{\hspace{1mm}}c@{\hspace{2mm}}c@{\hspace{1mm}}c@{\hspace{1mm}}c@{\hspace{1mm}}c@{}}
\bf Observed  & \bf Initializiat.  &\bf GBO & \bf Simplex & \bf LiDO & \bf LiDO-R  \\

    \expandafter\includegraphics\expandafter[width=0.16\textwidth,trim = 10px 30px 0px 20px, clip]{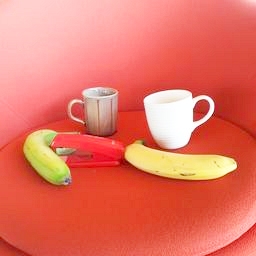} & 
  \begin{overpic}[width=0.16\textwidth,trim = 10px 30px 0px 20px, clip]{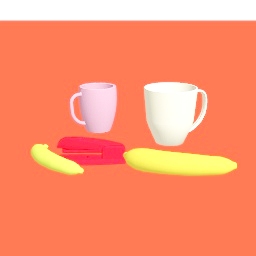}
     \put(0,0){\includegraphics[width=0.16\textwidth,trim = 10px 30px 0px 20px, clip]{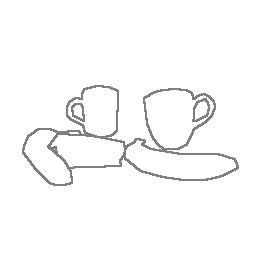}}
  \end{overpic}&

  \begin{overpic}[width=0.16\textwidth,trim = 10px 30px 0px 20px, clip]{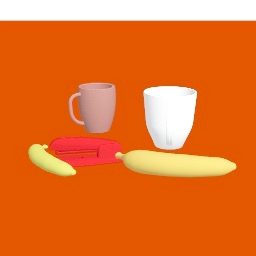}
     \put(0,0){\includegraphics[width=0.16\textwidth,trim = 10px 30px 0px 20px, clip]{img/imgs_real/76CO}}
  \end{overpic}&
    \begin{overpic}[width=0.16\textwidth,trim = 10px 30px 0px 20px, clip]{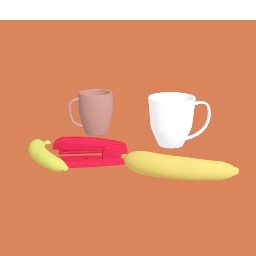}
     \put(0,0){\includegraphics[width=0.16\textwidth,trim = 10px 30px 0px 20px, clip]{img/imgs_real/76CO}}
  \end{overpic}&
  \begin{overpic}[width=0.16\textwidth,trim = 10px 30px 0px 20px, clip]{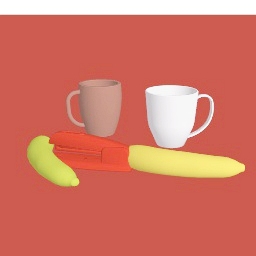}
     \put(0,0){\includegraphics[width=0.16\textwidth,trim = 10px 30px 0px 20px, clip]{img/imgs_real/76CO}}
  \end{overpic}&
    \begin{overpic}[width=0.16\textwidth,trim = 10px 30px 0px 20px, clip]{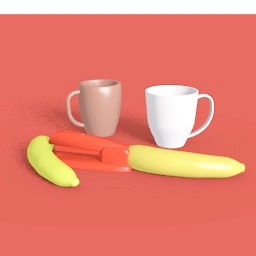}
  \end{overpic}\\

\multicolumn{6}{p{0.98\textwidth}}{ \footnotesize{ 
The left-hand banana size and pose is wrong in the Init, only  \ourMeth fits it properly, overall good performance of all the methods, e.g.\ the left-hand mug obtains brown colour.
}}\\
  \\
  
    \expandafter\includegraphics\expandafter[\www,trim = 0px 20px 0px 10px,clip]{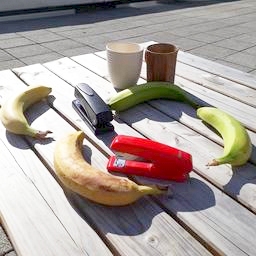} & 
  \begin{overpic}[width=0.16\textwidth,trim = 0px 20px 0px 10px,clip]{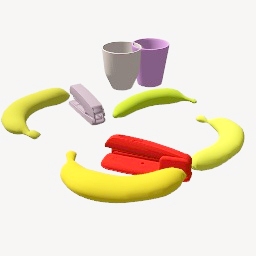}
     \put(0,0){\includegraphics[width=0.16\textwidth,trim = 0px 20px 0px 10px,clip]{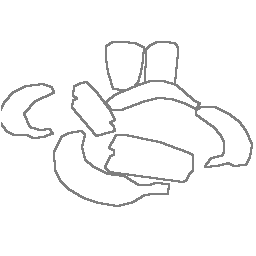}}
  \end{overpic}&

  \begin{overpic}[width=0.16\textwidth,trim = 0px 20px 0px 10px,clip]{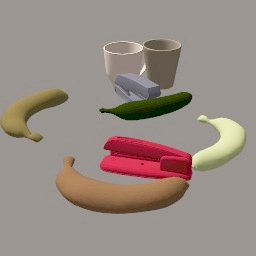}
     \put(0,0){\includegraphics[width=0.16\textwidth]{img/imgs_real/95CO}}
  \end{overpic}&
    \begin{overpic}[width=0.16\textwidth,trim = 0px 20px 0px 10px,clip]{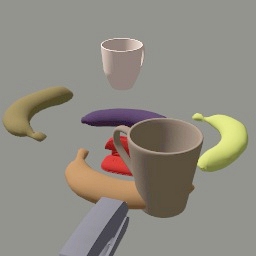}
     \put(0,0){\includegraphics[width=0.16\textwidth,trim = 0px 20px 0px 10px,clip]{img/imgs_real/95CO}}
  \end{overpic}&
  \begin{overpic}[width=0.16\textwidth,trim = 0px 20px 0px 10px,clip]{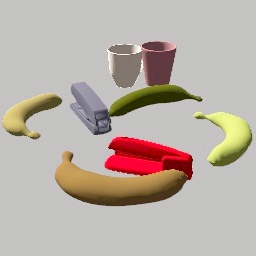}
     \put(0,0){\includegraphics[width=0.16\textwidth,trim = 0px 20px 0px 10px,clip]{img/imgs_real/95CO}}
  \end{overpic}&
    \begin{overpic}[width=0.16\textwidth,trim = 0px 20px 0px 10px,clip]{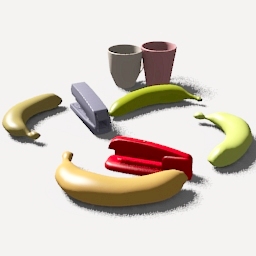}
  \end{overpic}
  \\
    \multicolumn{6}{p{0.98\textwidth}}{ \footnotesize{ 
Difficult scene, here GBO and Simp diverge objects, \ourMeth works well (see the gray stapler and the banana near the mugs).  
  }}\\
\\
          \expandafter\includegraphics\expandafter[clip,width=0.16\textwidth,trim =0px 20px 0px 10px]{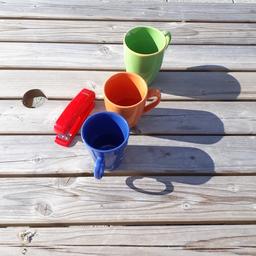}  & 
  \begin{overpic}[width=0.16\textwidth,trim = 0px 20px 0px 10px,clip]{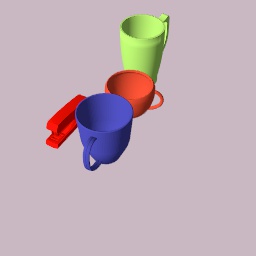}
     \put(0,0){\includegraphics[width=0.16\textwidth,trim = 0px 20px 0px 10px,clip]{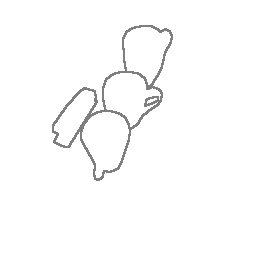}}
  \end{overpic}&

  \begin{overpic}[width=0.16\textwidth,trim = 0px 20px 0px 10px,clip]{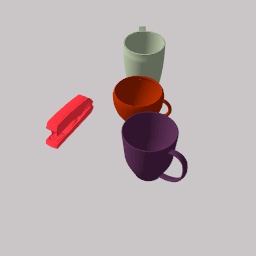}
     \put(0,0){\includegraphics[width=0.16\textwidth,trim = 0px 20px 0px 10px,clip]{img/imgs_real/97CO}}
  \end{overpic}&
    \begin{overpic}[width=0.16\textwidth,trim = 0px 20px 0px 10px,clip]{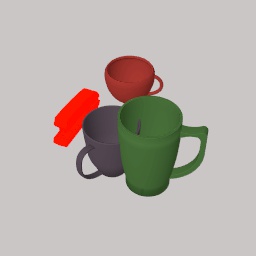}
     \put(0,0){\includegraphics[width=0.16\textwidth,trim = 0px 20px 0px 10px,clip]{img/imgs_real/97CO}}
  \end{overpic}&
  \begin{overpic}[width=0.16\textwidth,trim = 0px 20px 0px 10px,clip]{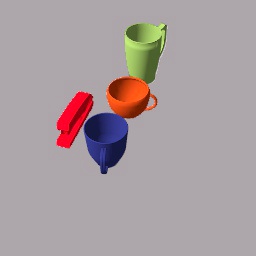}
     \put(0,0){\includegraphics[width=0.16\textwidth,trim = 0px 20px 0px 10px,clip]{img/imgs_real/97CO}}
  \end{overpic}&
    \begin{overpic}[width=0.16\textwidth,trim = 0px 20px 0px 10px,clip]{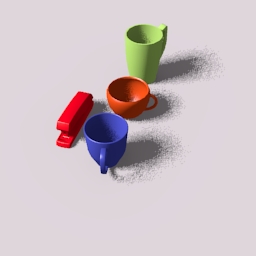}
  \end{overpic}
  \\
    \multicolumn{6}{p{0.98\textwidth}}{ \footnotesize{ 
GBO and Simplex corrupt the initialization, \ourMeth improves the object poses.
  }}\\
  \\
    \expandafter\includegraphics\expandafter[clip,width=0.16\textwidth,trim =0px 20px 0px 10px]{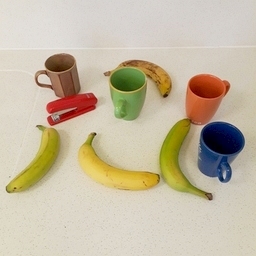} &  
  \begin{overpic}[width=0.16\textwidth,trim = 0px 20px 0px 10px, clip]{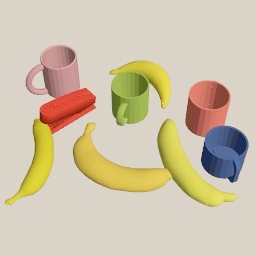}
     \put(0,0){\includegraphics[width=0.16\textwidth,trim = 0px 20px 0px 10px,clip]{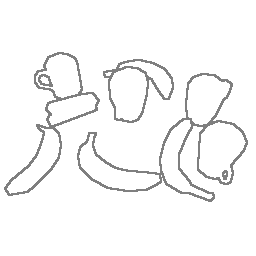}}
  \end{overpic}&

  \begin{overpic}[width=0.16\textwidth,trim = 0px 20px 0px 10px,clip]{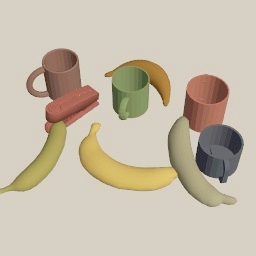}
     \put(0,0){\includegraphics[width=0.16\textwidth,trim = 0px 20px 0px 10px,clip]{img/imgs_real/114CO}}
  \end{overpic}&
    \begin{overpic}[width=0.16\textwidth,trim = 0px 20px 0px 10px,clip]{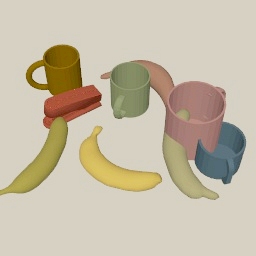}
     \put(0,0){\includegraphics[width=0.16\textwidth,trim = 0px 20px 0px 10px,clip]{img/imgs_real/114CO}}
  \end{overpic}&
  \begin{overpic}[width=0.16\textwidth,trim = 0px 20px 0px 10px,clip]{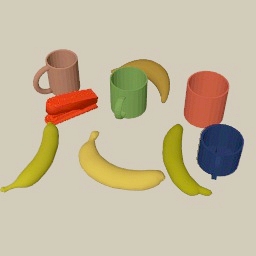}
     \put(0,0){\includegraphics[width=0.16\textwidth,trim = 0px 20px 0px 10px,clip]{img/imgs_real/114CO}}
  \end{overpic}&
    \begin{overpic}[width=0.16\textwidth,trim = 0px 20px 0px 10px,clip]{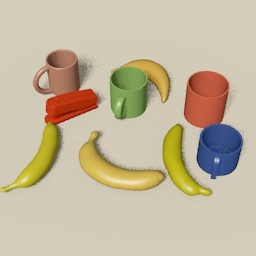}
  \end{overpic}
 \\ 
    \multicolumn{6}{p{0.98\textwidth}}{ \footnotesize{ 
All the methods update the colours and poses of most of the objects, yet \ourMeth is much more accurate (for example, compare each of the four bananas). The colours of \ourMeth of all the objects are well predicted (compare output of each method to the observed image). }}\\

\end{tabular}
  \caption{Example runs for Real dataset, the order is the same as in
    Figure~\ref{i:examp}. Also note that for each image the camera viewpoint is initialized accurately, and how similar the Observed and
    LiDO-R images are.
Obtaining an exact match to the ground truth outline may be impossible because we only have a fixed set of shapes to choose from, none of which may match the actual object shape.    
    }\label{i:exampreal}
  \end{center}
\end{figure*}

Note the initialized CAD shapes for real images are well matched (see
similar object shapes in Figure~\ref{i:exampreal}), even though these
shapes were never observed during training. The objects and global LVs
are then refined well. This was facilitated by introducing shape mismatch
in the second noisy dataset source of LiDO (see Section~\ref{sec:expts_trainingdata}).

The results in Tables~\ref{t:resSPIX} and \ref{t:resRPIX} afford
a direct comparison of the optimizers, all using the OpenGL renderer.
However, we can also run \ourMeth with its ``native'' renderer
Blender (shown as the LiDO-R column in Figures~\ref{i:examp} and
\ref{i:exampreal}).
We calculated the MSE errors for Blender renderer (IoUs are the same
for both renderers since only the appearance changes). These MSE errors were
similar to  LiDO that used OpenGL renderer 
(for Synthetic dataset: 21.6 [ob] and 11.3 [gr], for Real
dataset: 40.9 [ob] and 23.0 [gr]).

Figure~\ref{i:exampreal}
shows example runs and explanatory text for real image examples. 
In general  \ourMeth  obtains better results in a shorter
test-time than the alternatives, and usually converges to a better
configuration. \ourMeth also has the advantage that it can be trained
to handle model mismatch, as shown in the real dataset experiments. More examples are given
 in \ref{a:morereal}.

\section{Discussion}
\label{s:ResConc}

Above we have
demonstrated LiDO, a full framework for the initialization and refinement of a 3D
representation of the scene from a single image.
 The main features of
LiDO are: the advantage of not requiring an error metric $E$ to be
defined in image space, rapid convergence, and robust refinement in 
the presence of noise and distractors. LiDO is generally robust
 to issues that are common for error-based methods: 
 the updates can point in wrong direction when dealing
with cluttered scenes and shadows in observed images; difficulties can
arise from an inability to exactly match the target object with one of
a different shape; and when predicted objects overlap the background or
other objects. LiDO is generally robust to such problems as it  directly
learns to optimize in the latent space. Our method is not limited to 
rigid objects, one could use LiDO for \eg  
multiple-human pose estimation, or hand-pose and appearance reconstruction.

One apparent limitation of LiDO is that we need to train an additional 
 Prediction Network  in advance  for a particular dataset. 
Incorporating neural networks requires extra time for training, 
but allows for smarter LV updates.

Another potential limitation is a need for a synthetic training dataset.
 Although such a dataset is required to train the LiDO Prediction Network,
for any method one also needs to use an initializer of the scene graph LVs. This 
 means that: i) the initializer was trained already on such dataset, 
and ii) a 3D graphics representation of a scene graph LVs would be available. 
This representation and the dataset can be reused for LiDO, 
e.g.\ by computing the initialization errors, 
or by adding noise to the ground truth LVs.
The advantages of LiDO mean that it could be a critical component 
in the development of future vision-as-inverse-graphics systems.

\subsection*{Acknowledgments}
We thank the anonymous referees for their comments which have helped
improve the paper.

LR was supported by Microsoft Research through its PhD Scholarship
Programme. The work of CW was supported in part by The Alan Turing
Institute under the EPSRC grant EP/N510129/1.

{\small
\bibliographystyle{ieee}
\bibliography{bib}
}
\clearpage
\appendix
 
\begin{figure*}
\section{More Examples of Prediction for Synthetic Dataset}
\label{a:moresyn}
\edef\www{width=0.16\textwidth}
\renewcommand\arraystretch{0.7}
\begin{tabular}{@{}c@{\hspace{1mm}}c@{\hspace{2mm}}c@{\hspace{1mm}}c@{\hspace{1mm}}c@{\hspace{1mm}}c@{}}
\bf Observed  & \bf Initializiat.  &\bf GBO & \bf Simplex & \bf LiDO & \bf LiDO-R  \\

\expandafter\includegraphics\expandafter[\www]{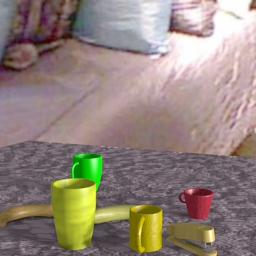} & 
  \frame{\begin{overpic}[width=0.16\textwidth]{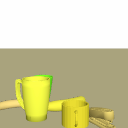}
     \put(0,0){\includegraphics[width=0.16\textwidth]{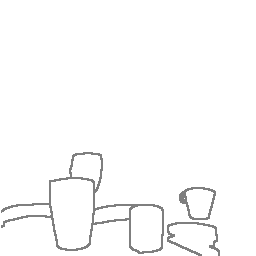}}
  \end{overpic}}&

  \frame{\begin{overpic}[width=0.16\textwidth]{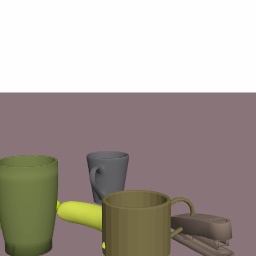}
     \put(0,0){\includegraphics[width=0.16\textwidth]{img/imgs/248CO}}
  \end{overpic}}&
    \frame{\begin{overpic}[width=0.16\textwidth]{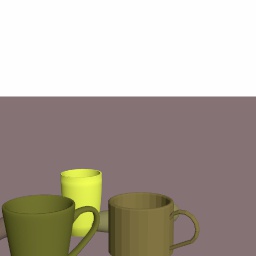}
     \put(0,0){\includegraphics[width=0.16\textwidth]{img/imgs/248CO}}
  \end{overpic}}&
  \frame{\begin{overpic}[width=0.16\textwidth]{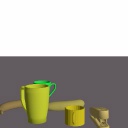}
     \put(0,0){\includegraphics[width=0.16\textwidth]{img/imgs/248CO}}
  \end{overpic}}&
    \frame{\begin{overpic}[width=0.16\textwidth]{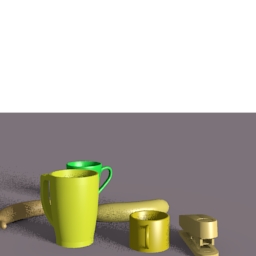}
  \end{overpic}}
  
  \\

\multicolumn{6}{p{0.98\textwidth}}{   \footnotesize{   {\bf {Poor initialization}}: GBO and Simplex converge to wrong configurations of
object poses and colours, while \ourMeth is robust to
initialization errors; note here the initialized object sizes  are wrong and \ourMeth improves all the detected objects.    }} \\

  \\

\expandafter\includegraphics\expandafter[\www]{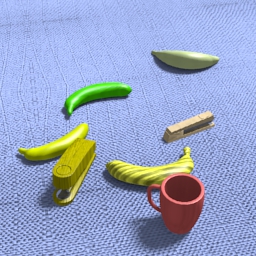} & 
  \begin{overpic}[width=0.16\textwidth]{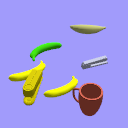}
     \put(0,0){\includegraphics[width=0.16\textwidth]{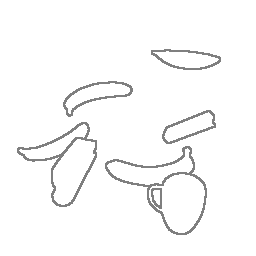}}
  \end{overpic}&

  \begin{overpic}[width=0.16\textwidth]{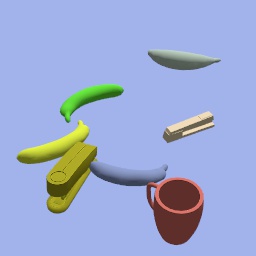}
     \put(0,0){\includegraphics[width=0.16\textwidth]{img/imgs/246CO}}
  \end{overpic}&
    \begin{overpic}[width=0.16\textwidth]{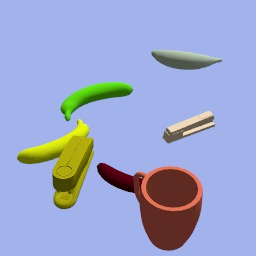}
     \put(0,0){\includegraphics[width=0.16\textwidth]{img/imgs/246CO}}
  \end{overpic}&
  \begin{overpic}[width=0.16\textwidth]{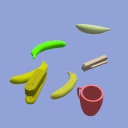}
     \put(0,0){\includegraphics[width=0.16\textwidth]{img/imgs/246CO}}
  \end{overpic}&
    \begin{overpic}[width=0.16\textwidth]{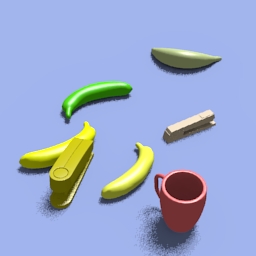}
  \end{overpic}
  
  \\

\multicolumn{6}{p{0.98\textwidth}}{  \footnotesize{   {\bf{Typical input (1)}}: 
There are 7 objects, 6 object converge properly for all the methods, the initialized position of the bottom banana in wrong, all the methods fail to fix it: GBO and Simplex corrupt the colour, LiDO maintains the yellow colour.}} \\
\\

\expandafter\includegraphics\expandafter[\www]{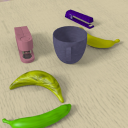} & 
  \begin{overpic}[width=0.16\textwidth]{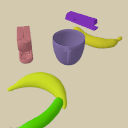}
     \put(0,0){\includegraphics[width=0.16\textwidth]{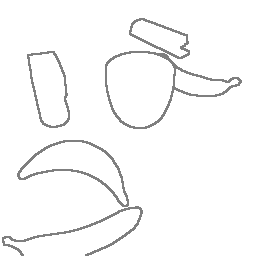}}
  \end{overpic}&

  \begin{overpic}[width=0.16\textwidth]{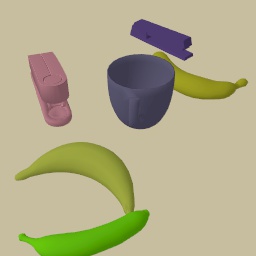}
     \put(0,0){\includegraphics[width=0.16\textwidth]{img/imgs/273CO}}
  \end{overpic}&
    \begin{overpic}[width=0.16\textwidth]{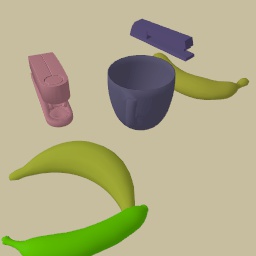}
     \put(0,0){\includegraphics[width=0.16\textwidth]{img/imgs/273CO}}
  \end{overpic}&
  \begin{overpic}[width=0.16\textwidth]{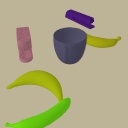}
     \put(0,0){\includegraphics[width=0.16\textwidth]{img/imgs/273CO}}
  \end{overpic}&
    \begin{overpic}[width=0.16\textwidth]{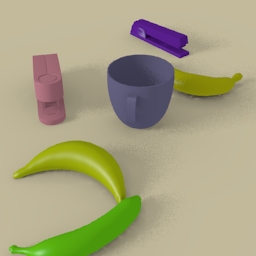}
  \end{overpic}
  
  \\

\multicolumn{6}{p{0.98\textwidth}}{  \footnotesize{{\bf{Typical input (2)}}: 
All objects are initialized well and converge properly, except the green banana for which the azimuthal rotation is wrong, all the methods improve the pose. Note well predicted shadows in LiDO-R.}} \\

\\

\expandafter\includegraphics\expandafter[\www]{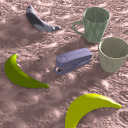} & 
  \begin{overpic}[width=0.16\textwidth]{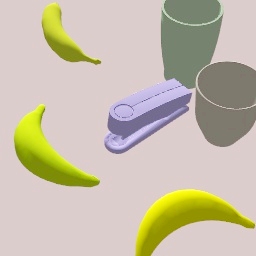}
     \put(0,0){\includegraphics[width=0.16\textwidth]{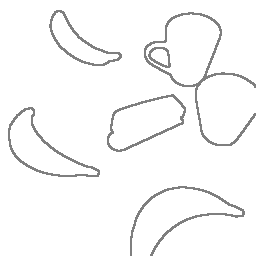}}
  \end{overpic}&

  \begin{overpic}[width=0.16\textwidth]{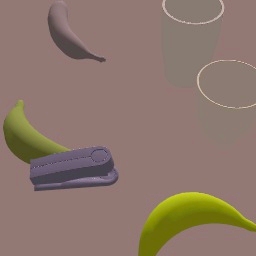}
     \put(0,0){\includegraphics[width=0.16\textwidth]{img/imgs/378CO}}
  \end{overpic}&
    \begin{overpic}[width=0.16\textwidth]{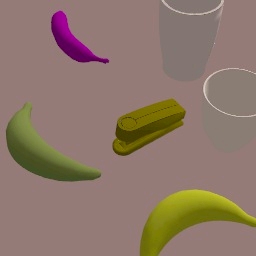}
     \put(0,0){\includegraphics[width=0.16\textwidth]{img/imgs/378CO}}
  \end{overpic}&
  \begin{overpic}[width=0.16\textwidth]{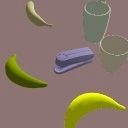}
     \put(0,0){\includegraphics[width=0.16\textwidth]{img/imgs/378CO}}
  \end{overpic}&
    \begin{overpic}[width=0.16\textwidth]{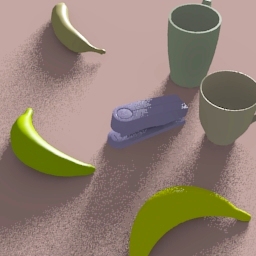}
  \end{overpic}
  
  \\

\multicolumn{6}{p{0.98\textwidth}}{  \footnotesize{ {\bf{Strong textures and shadows}}: 
For GBO  the blue stapler (middle) diverges, for Simplex it becomes brown, 
\ourMeth is robust to
such distractors: stapler pose/size improves, both mugs become smaller with proper colours (also compare Observed and LiDO-R).}} \\

\end{tabular}
  \caption{Example runs for Synthetic dataset, showing from left: the observed input image, the
    initialization (OpenGL), and images after refinement for GBO, Simplex and LiDO using OpenGL renderers, and LiDO using Blender renderer (LiDO-R). We overlay black contours of the ground truth object masks on top of each OpenGL image to ease the comparison of object poses. 
    \vspace{3cm}
    \label{i:exampAPP}}
    
\end{figure*}

\begin{figure*}[h!t!]
\section{More Examples of Prediction for Real Dataset}
\label{a:morereal}
\begin{center}
\edef\www{width=0.16\textwidth}
\renewcommand\arraystretch{0.5}
\begin{tabular}{@{}c@{\hspace{1mm}}c@{\hspace{2mm}}c@{\hspace{1mm}}c@{\hspace{1mm}}c@{\hspace{1mm}}c@{}}
\bf Observed  & \bf Initializiat.  &\bf GBO & \bf Simplex & \bf LiDO & \bf LiDO-R  \\

      \expandafter\includegraphics\expandafter[\www]{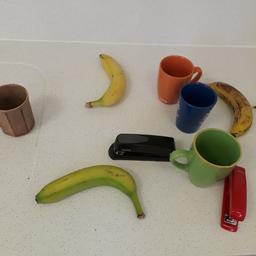} & 
  \begin{overpic}[width=0.16\textwidth]{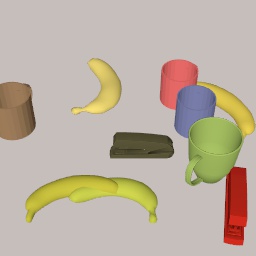}
     \put(0,0){\includegraphics[width=0.16\textwidth]{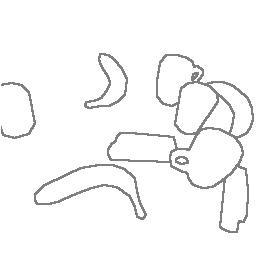}}
  \end{overpic}&

  \begin{overpic}[width=0.16\textwidth]{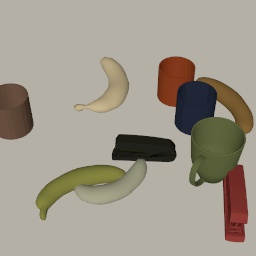}
     \put(0,0){\includegraphics[width=0.16\textwidth]{img/imgs_real/9CO}}
  \end{overpic}&
    \begin{overpic}[width=0.16\textwidth]{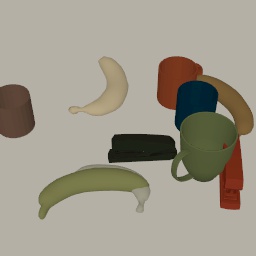}
     \put(0,0){\includegraphics[width=0.16\textwidth]{img/imgs_real/9CO}}
  \end{overpic}&
  \begin{overpic}[width=0.16\textwidth]{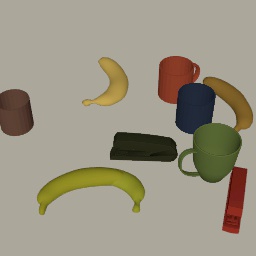}
     \put(0,0){\includegraphics[width=0.16\textwidth]{img/imgs_real/9CO}}
  \end{overpic}&
    \begin{overpic}[width=0.16\textwidth]{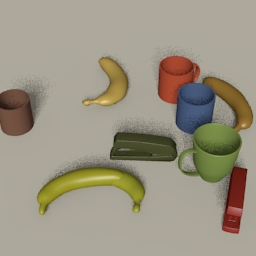}
  \end{overpic}
  \\

  \multicolumn{6}{p{0.98\textwidth}}{ \footnotesize{ 
All methods improve the colours, note double detection of the front banana (for the Observed banana in the front, in the Initialization there are two bananas intersecting each other) and different behaviours for this object.\newline
  }}\\

      \expandafter\includegraphics\expandafter[\www]{img/imgs_real/77gt} & 
  \begin{overpic}[width=0.16\textwidth]{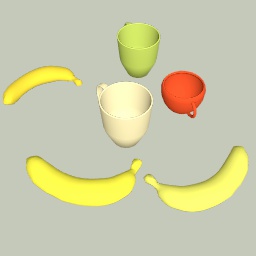}
     \put(0,0){\includegraphics[width=0.16\textwidth]{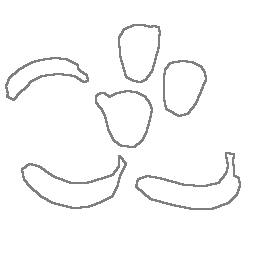}}
  \end{overpic}&

  \begin{overpic}[width=0.16\textwidth]{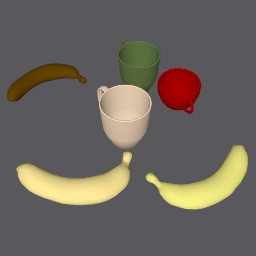}
     \put(0,0){\includegraphics[width=0.16\textwidth]{img/imgs_real/77CO}}
  \end{overpic}&
    \begin{overpic}[width=0.16\textwidth]{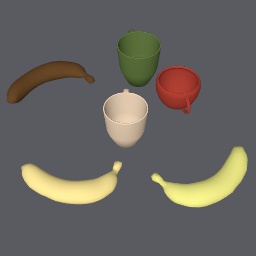}
     \put(0,0){\includegraphics[width=0.16\textwidth]{img/imgs_real/77CO}}
  \end{overpic}&
  \begin{overpic}[width=0.16\textwidth]{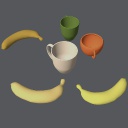}
     \put(0,0){\includegraphics[width=0.16\textwidth]{img/imgs_real/77CO}}
  \end{overpic}&
    \begin{overpic}[width=0.16\textwidth]{img/imgs_real/77l}
  \end{overpic}
  \\

  \multicolumn{6}{p{0.98\textwidth}}{ \footnotesize{ 
All the methods improve the ground plane colour. None of the methods perform well on the switched-orientation banana on the right.\newline
  }}\\

       \expandafter\includegraphics\expandafter[\www]{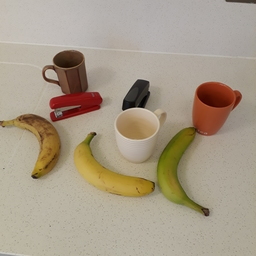} & 
  \begin{overpic}[width=0.16\textwidth]{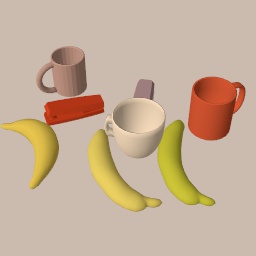}
     \put(0,0){\includegraphics[width=0.16\textwidth]{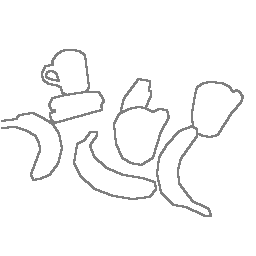}}
  \end{overpic}&

  \begin{overpic}[width=0.16\textwidth]{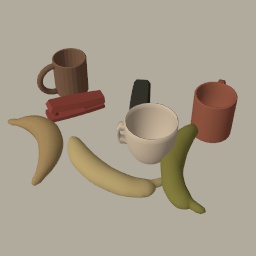}
     \put(0,0){\includegraphics[width=0.16\textwidth]{img/imgs_real/120CO}}
  \end{overpic}&
    \begin{overpic}[width=0.16\textwidth]{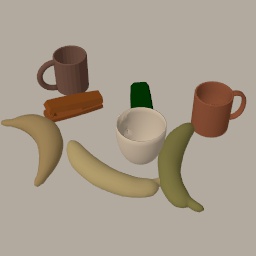}
     \put(0,0){\includegraphics[width=0.16\textwidth]{img/imgs_real/120CO}}
  \end{overpic}&
  \begin{overpic}[width=0.16\textwidth]{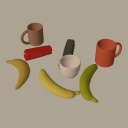}
     \put(0,0){\includegraphics[width=0.16\textwidth]{img/imgs_real/120CO}}
  \end{overpic}&
    \begin{overpic}[width=0.16\textwidth]{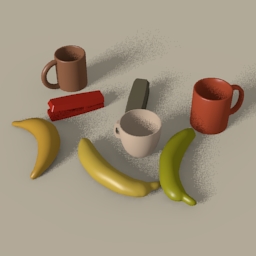}
  \end{overpic}
  \\
    \multicolumn{6}{p{0.98\textwidth}}{ \footnotesize{ 
All the methods improve the object poses and colours, note the refinement behaviour of the occluded black stapler.\newline
  }}\\

       \expandafter\includegraphics\expandafter[\www]{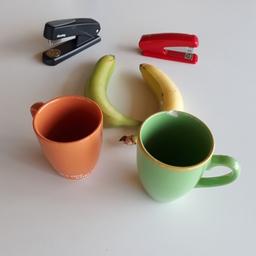} & 
  \begin{overpic}[width=0.16\textwidth]{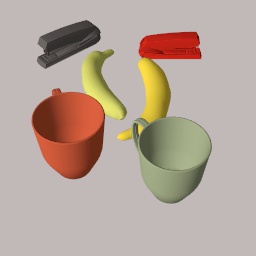}
     \put(0,0){\includegraphics[width=0.16\textwidth]{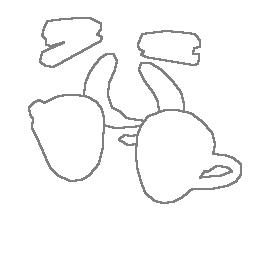}}
  \end{overpic}&

  \begin{overpic}[width=0.16\textwidth]{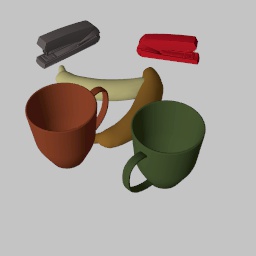}
     \put(0,0){\includegraphics[width=0.16\textwidth]{img/imgs_real/28CO}}
  \end{overpic}&
    \begin{overpic}[width=0.16\textwidth]{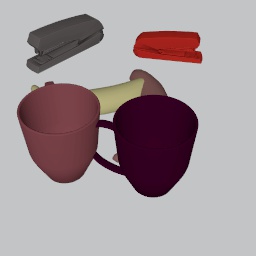}
     \put(0,0){\includegraphics[width=0.16\textwidth]{img/imgs_real/28CO}}
  \end{overpic}&
  \begin{overpic}[width=0.16\textwidth]{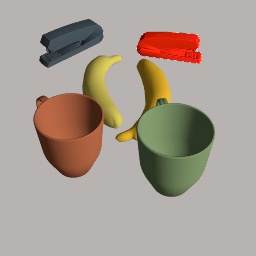}
     \put(0,0){\includegraphics[width=0.16\textwidth]{img/imgs_real/28CO}}
  \end{overpic}&
    \begin{overpic}[width=0.16\textwidth]{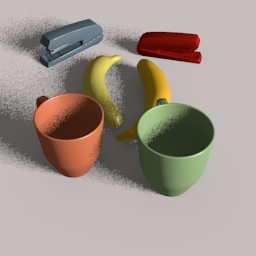}
  \end{overpic}
  \\
    \multicolumn{6}{p{0.98\textwidth}}{ \footnotesize{ 
GBO and Simplex make the Init worse: bananas are rotated, mugs have wrong colours, \ourMeth improves the colours of mugs. Only LiDO refines the handle of the orange mug, and none of the methods correctly identify the handle position of the green mug.
  }}\\

 \end{tabular}
  \caption{Example runs for Real dataset, the order is the same as in
    Figure~\ref{i:exampAPP}. Also note how similar the Observed and
    LiDO-R images are.}\label{i:examprealAPP}
  \end{center}
      \vspace{4cm}
\end{figure*}

\clearpage

\section{CNN Architectures of Detector and LV Initialization Networks}
\label{a:det}

\renewcommand{\arraystretch}{1.01}
\begin{table*}[h!t!]
\centering

\begin{tabular}{ c c }\toprule
\multicolumn{2}{c}{\bf Detector}  \\
 Class & Size   \\
\cmidrule{1-2}
\multicolumn{2}{c}{Input $128 \times 128 \times 3$ (Image)}\\
\cmidrule{1-2}
 \multicolumn{2}{c}{  VGG-16 (all 13 convolutional layers)}\\
\cmidrule{1-2}
\multicolumn{2}{c}{ 3 $\times$ C-50-6 \emph{(separate per network)}} \\
\cmidrule{1-2}
Fd-200 & Fd-200 \\
Softmax-4 & Sigm-1 \\
\bottomrule
\end{tabular}

\vspace{0.5cm}

\begin{tabular}{  c c  c c }\toprule

\multicolumn{4}{c}{\bf LV Initialization Networks} \\

 Shape &  Azimuth (Ob/Lighting) & Lighting & Camera   \\
\cmidrule{1-4}
\multicolumn{4}{c}{Input $128 \times 128 \times 3$ (Image)}\\
\cmidrule{1-4}
 \multicolumn{4}{c}{  VGG-16 (all 13 convolutional layers)}\\
\cmidrule{1-4}
\multicolumn{4}{c}{  3 $\times$  C-50-6 \emph{(separate per network)}} \\
\cmidrule{1-4}
 Fd-50 & Fd-50 & Fd-100 & Fd-50\\
Softmax-6/15/8 & Softmax-18 & Sigm-3 & Sigm-2\\
\bottomrule
\end{tabular}
\vspace{0.5cm}

\begin{tabular}{ c c c c  c c }\toprule
\multicolumn{6}{c}{{\bf Learning rates}} \\
\midrule
 Class & Size & Shape & Azimuth & Lighting & Camera   \\
0.001 & 0.0002 & 0.001 & 0.0003 & 0.0001 & 0.0001\\
\bottomrule
\end{tabular}
\caption{The configurations of the detector (top) and LV initialization networks (middle), and the learning rates (bottom).  Layer description  (where N denotes the number of units and K the filter size), is as follows: 
  1) \emph{Convolutional layer}: C-N-K;
  2) \emph{Fully connected layer, with its input concatenated with the detector output (position of the detection plus object size)}: Fd-N;
 3) \emph{Sigmoid (fully connected) layer}: Sigm-N;
 4) \emph{Softmax (fully connected) layer}: Softmax-N. Colour networks (for objects and for ground plane) are simple 3-layer CNNs with leaky rectify activations: Input, C-27-6 (stride 6, dropout $p=0.5$), Fd-40, F-40, Sigm-3; trained with 0.0001 learning rate.
}
\label{t:layers}
\end{table*}

The detector is trained on 30,000 positive patches with objects central contact point centred (with small noise of $\pm$ 8 pixels added), and 90,000 negative patches (30,000 random patches, 30,000 patches with the centre nearby the central contact point of other objects, and 30,000 random crops from the ImageNet\ \cite{ILSVRC15} dataset). The detector is run on  10,000 images to produce the training dataset for the initialization networks. Afterwards, we  apply the LV initialization networks on another 10,000 images to produce the dataset for LiDO (the first source).

Table~\ref{t:layers} shows the network configurations and learning rates used for training. We use all 13
convolutional layers of VGG-16 as the core on 128 $\times$ 128 pixel input. We use only the first three pooling layers, and the VGG weights are kept fixed. 
Since the original pixel values are integers in $[0,255]$, while the VGG expects zero mean pixel intensity, we subtract the mean. The region outside the image frame is given as value 0.

VGG layer activations are \emph{ReLU}, layers on top of VGG use \emph{tanh} activations.
We do not use padding in our VGG layers. The fully
connected layers of the detector networks are implemented as filter $1 \times
1$ convolutional layers, so they can be efficiently applied in a sliding window
manner.

 The implementation is in Python (Theano) and we use the Adam~\cite{KingmaB14} optimizer with L2 or categorical cross-entropy loss to train the networks.  Each LV belongs to a specific LV-set responsible for a given property, and we train one LV initialization network per global LV-set and one LV initialization network per each object class for object LVs. We use dropout in the detector networks with $p=0.5$ in all the 3 convolutional layers (on top of VGG ones), and after the first one for the initialization networks.

\end{document}